\documentclass[10pt,twocolumn,letterpaper]{article}

\usepackage{cvpr}
\usepackage{times}
\usepackage{epsfig}
\usepackage{graphicx}
\usepackage{amsmath}
\usepackage{amssymb}
\newtheorem{theorem}{Theorem}
\numberwithin{theorem}{section}

\newtheorem{lemma}[theorem]{Lemma}

\newtheorem{result}[theorem]{Result}

\newcommand{\C}{{\mathbb C}}

\newcommand{\M}[1]{\mathtt{#1}}
\newcommand{\V}[1]{\mathbf{#1}}
\newcommand{\arr}[2]{\begin{array}{#1} #2\end{array}}
\newcommand{\mat}[2]{\left[\!\!\arr{#1}{#2}\!\!\right]}
\newcommand{\xx}[1]{\left[#1\right]_{\times}}

\newcommand{\comment}[1]{}
\newcommand{\Figs}{Figures}
\def\gb{Gr{\"o}bner basis\xspace}

\def\gbs{Gr{\"o}bner bases\xspace}

\def\ntm#1#2{\begingroup\setlength{\medmuskip}{0mu}$\textrm{#1}\times\textrm{#2}$\endgroup}

\def\fFf{$\textrm{f+E+f}$\xspace}
\def\Ef{$\textrm{E+f}$\xspace}
\def\Efk{$\textrm{E+f+k}$\xspace}
\newcommand{\myparagraph}[1]{\vspace*{-10pt}\paragraph{#1}\mbox{}\\}

\usepackage[pagebackref=true,breaklinks=true,colorlinks,bookmarks=false]{hyperref}
\cvprfinalcopy 
\def\cvprPaperID{2027} 

\ifcvprfinal\fi
\begin{document}

\title{A clever elimination strategy for efficient minimal solvers}
\author{%
Zuzana Kukelova\\
Czech Technical University in Prague, Czechia\\
{\tt\small kukelzuz@fel.cvut.cz}
\and
Joe Kileel\\
University of California, Berkeley, USA\\
{\tt\small jkileel@math.berkeley.edu}
\and
Bernd Sturmfels\\
University of California, Berkeley, USA\\
{\tt\small bernd@math.berkeley.edu}
\and
Tomas Pajdla\\
Czech Technical University in Prague, Czechia\\
{\tt\small pajdla@cvut.cz}}
\maketitle
\begin{abstract}
\noindent 
We present a new insight into the systematic generation of minimal solvers in computer vision,
which leads to smaller and faster solvers. Many minimal problem formulations 
are coupled sets of linear and polynomial equations where image measurements enter the linear equations only. We show that it is useful to solve such systems by first eliminating all the unknowns that do not appear in the linear equations and then extending solutions to the rest of unknowns. This can be generalized to fully non-linear systems by linearization via lifting. We demonstrate that this approach leads to more efficient solvers in three problems of partially calibrated relative camera pose computation with unknown focal length and/or radial distortion. Our approach also generates new interesting constraints on the fundamental matrices of partially calibrated cameras, which were not known before.   
\end{abstract}

\section{Introduction}
\noindent Computing multi-view geometry is one of the most basic and important tasks in computer vision~\cite{HZ-2003}. These include minimal problem solvers~\cite{Nister-5pt-PAMI-2004,Kukelova-ECCV-2008} in, e.g., structure from motion~\cite{Snavely-IJCV-2008}, visual navigation~\cite{DBLP:journals/ram/ScaramuzzaF11}, large scale 3D reconstruction~\cite{DBLP:conf/cvpr/HeinlySDF15} and image based localization~\cite{Sattler16PAMI}. Fast, and efficient, minimal solvers are instrumental in RANSAC~\cite{Fischler-Bolles-ACM-1981} based robust estimation algorithms~\cite{DBLP:journals/pami/RaguramCPMF13}. 
\begin{figure}[t]
\label{fig:teaser}
\begin{center}
\includegraphics[width=0.8\linewidth]{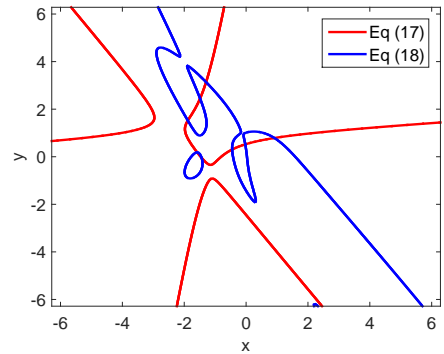}
\caption{An illustration of the two equations~\eqref{eq:cubic} and~\eqref{eq:quintic}, which define the \fFf problem, cut by six linear equations for six image point correspondences.}
\end{center}
\end{figure}

In this paper we present a new insight into the systematic generation of minimal solvers~\cite{Kukelova-ECCV-2008}, which leads to smaller and faster solvers. We explain our approach 
in the context of elimination theory~\cite{Cox-IVA-2015} and we offer an
 interpretation of the theory that is useful for practice in computer vision.

Our main technical contribution is  a new strategy for solving minimal problems.
For many computer vision applications, that strategy
allows to do more computation in an off-line stage and less computation in an on-line stage.

We exploit that many minimal problems in
 computer vision lead to coupled sets of linear and polynomial equations where image measurements enter the linear equations only. We show how to solve such systems efficiently by first eliminating all unknowns which do not appear in the linear equations, and then extending solutions to the other unknowns. Moreover, our approach can be generalized to fully non-linear systems by linearization via monomial lifting~\cite{DBLP:conf/eurocrypt/CourtoisKPS00}.

We demonstrate that this approach leads to more efficient on-line solvers in three problems of partially calibrated relative camera pose computation with unknown focal length and/or radial distortion.  
Interestingly, our approach also generates new constraints on the fundamental matrices of partially calibrated cameras, which were not known before.

\subsection{Related work}
\noindent Historically, minimal problems~\cite{Grunert-1841, Sturm-1869,Longuet-Higgins-Nature-1981,DBLP:journals/ivc/Longuet-Higgins92} addressed problems in geometry of one and two perspective cameras. Later, a more systematic approach to solving minimal problems in computer vision appeared, e.g., in~\cite{Nister-5pt-PAMI-2004,DBLP:journals/ijcv/NisterS06,DBLP:journals/jmiv/QuanTM06, DBLP:conf/cvpr/KukelovaP07,DBLP:journals/jmiv/NisterS07,DBLP:journals/ivc/SteweniusNKS08}.  It developed a number of ad-hoc, as well as, systematic tools for solving polynomial systems appearing in computer vision. These were later used and improved by many researchers, e.g., ~\cite{Byrod-ICCV-2007, Byrod-ECCV-2008, Kukelova-ECCV-2008,
DBLP:conf/cvpr/RamalingamS08,DBLP:conf/cvpr/KneipSS11, 
DBLP:conf/iccv/NaroditskyD11,Hartley-PAMI-2012, DBLP:conf/eccv/CamposecoSP16,DBLP:conf/eccv/KneipSP12, DBLP:conf/iccv/KukelovaHBFP15, DBLP:conf/iccv/HeschR11, Larsson2016}. Lately, the
algebraic geometry foundations for computer vision came into focus in
  {\em algebraic vision}~\cite{DBLP:journals/corr/abs-1107-2875,DBLP:journals/moc/AholtO14,DBLP:journals/corr/AgarwalLST15, DBLP:journals/ijac/JoswigKSW16, Trager-IJCV-2016}. 

One of the key elements in computer vision applications has been to design procedures for solving special polynomial systems that move the computation from the on-line 
stage of solving equations to an earlier off-line stage~\cite{Kukelova-thesis}. 

Interestingly,  elimination theory~\cite{Cox-IVA-2015} has not been fully exploited in
such computer vision applications, although it has been used in many works~\cite{Nister-5pt-PAMI-2004,DBLP:journals/ivc/SteweniusNKS08,Stewenius-ISPRS-2006,Kukelova-ECCV-2008,Bujnak-ICCV-2009} implicitly.
%
\subsection{The main idea}
\noindent Our main idea is to use elimination theory to do more computation in the off-line stage and less in the on-line stage.

Natural formulations of vision models often involve more unknowns than those that appear in the linear constraints depending on image measurements. For instance, the constraint $\det{\M{F}}=0$ in fundamental matrix computation does not involve any image measurements. We argue that it is advantageous to pre-process such models by computing its projection into the space of relevant unknowns. This is done by elimination. Solving the linear equations on the
resulting projected variety is then fast. Subsequently, the values for the other unknowns can be determined using the Extension Theorem~\cite{Cox-IVA-2015} from computer algebra.

\section{Solving polynomial systems by elimination}
\label{sec:elimIdeal}
\noindent A classical (textbook)  strategy  for solving systems of  polynomial equations is to use elimination theory~\cite{Sturmfels-CBMS-2002,Cox-IVA-2015,Becker-GB-2003}.  The
strategy  consists  of  two  main steps.   
\begin{enumerate}
\item First,  the  equations  are ``simplified'' by eliminating some unknowns  to get a set of equations from which  the remaining unknowns  can be computed.  This  provides a set of partial solutions. 
\item Next, the partial solutions are extended to full solutions by substituting  the partial solutions back into the original equations and solving for the remaining unknowns. We next explain different elimination strategies.
\end{enumerate}
\subsection{Elimination strategies} 
\label{sec:strategies}
\subsubsection{Standard textbook elimination strategy}
Standard (textbook) elimination 
is based on the Elimination Theorem~\cite{Cox-IVA-2015}, which we review in
Theorem~\ref{thm:ElimTheorem} of the Appendix.
It states that, for an ideal $I \subset \C[x_{1}, \ldots, x_{n}]$, we can
read off LEX \gbs for all elimination ideals $I_l = I \cap \C[x_{1}, \ldots, x_{n}]$ from a LEX \gb $G$ for $I$. Here the sequence of elimination ideals ends with $I_n = I \cap \C[x_n]$.
This is generated by one equation in the single unknown $x_n$, which is then easy to solve numerically. 

For a polynomial system  with  a finite number  of solutions in $\C^n$, it  is  always possible~\cite[p.~254--255]{Cox-IVA-2015} to extend partial solutions from $x_{l+1}, \ldots, x_n$ to  $x_{l}, \ldots, x_n$.  For this, we choose  a  single polynomial $g$ with the lowest degree among all the  univariate polynomials in $x_{l}$ after substituting the partial solution into the polynomials in $x_{l+1}, \ldots, x_n$. 
%
\subsubsection{Standard computer vision elimination strategy}
\label{sec:CV_strategy}
In the existing minimal solvers, several different strategies for eliminating unknowns from the input equations were applied. These strategies were usually dependent on the
specific problem and were derived manually. Here we describe one strategy that was used in the vast majority of existing minimal solvers~\cite{Nister-5pt-PAMI-2004,DBLP:journals/ivc/SteweniusNKS08,Stewenius-ISPRS-2006,Kukelova-ECCV-2008,Bujnak-thesis,Bujnak-ICCV-2009}.

Consider a system of $m$ polynomial equations
$$ \left\{f_1(x_1,\dots,x_n) =0,\dots,f_m(x_1,\dots,x_n)=0 \right\}$$  
in $n$ unknowns $X \! = \! \left\{x_1,\dots,x_n\right\}$.
We assume that
the set $F = \{f_1,\dots,f_m\}$ generates a zero dimensional ideal $I \subset \C[X]$, 
\ie the system $F$ has a finite number of solutions.

In this strategy the set $F$ is partitioned into two subsets:
\begin{eqnarray}
\label{eq:F1}
    F_L &=& \left\{f_i \in F \;| \; deg(f_i) = 1\right\}, \\
    \label{eq:F2}
    F_N &=& \left\{f_i \in F \;| \; deg(f_i) > 1\right\}.    
\end{eqnarray}
This means that $F_L$ contains the linear polynomials from $F$ and $F_N$ contains the polynomials of higher degrees.

The linear equations $F_L$ can be rewritten as $\M{M} X_L = 0$,  where $X_L$ is a vector of all unknowns that appear in these equations. Then, the null space basis $\M{N}$ of $\M{M}$, i.e.\ $\M{M}\,\M{N}=0$, is used to parametrize the unknowns $X_L$ with new unknowns $Y$ via $X_L \!= \!\M{N}\,Y$.
 The parameterization $X_L \!=\! \M{N}\,Y$~is then plugged in the non-linear equations $F_N$. The system $F_N(Y \cup (X \setminus X_L))\!=\!0$ is solved using, e.g., a \gb method and the automatic generator of efficient solvers~\cite{Kukelova-ECCV-2008}. 
The solutions $Y$ are used to recover solutions $X_L \!=\! \M{N}\,Y$.
\subsubsection{A clever computer vision elimination strategy}
\label{sec:method}
\noindent
In computer vision, we often encounter polynomial systems in which only the linear equations $F_L$ depend on image measurements, while the nonlinear equations $F_N$ stay the same, regardless of image measurements. For example, the epipolar constraint~\cite{HZ-2003} generates linear equations that depend on the input measurements while the singularity of the fundamental matrix~\cite{HZ-2003} results in the non-linear equation $\det(\M{F}) = 0$, which does not depend on the measurements.

Here we present a new ``clever'' elimination strategy, which usually allows us to do more computation in the off-line stage and less computation in the on-line stage.

Throughout this section we assume that the nonlinear equations $F_N$ do not depend on image measurements, \ie for all instances these equations are the same. Later, in Section~\ref{sec:Efk}, we will show  how to deal with problems when $F_N$ contains equations that depend on image measurements.

Let us now describe our new elimination strategy. We first divide the input equations $F$ into the linear equations $F_L$ in
\eqref{eq:F1} and the non-linear equations $F_N$ in \eqref{eq:F2}.
Moreover, we divide the $n$ given unknowns $X$ into two subsets:
\begin{eqnarray}
\label{eq:X1}
    X_L &=& \left\{x_i \in X \,|\, x_i\textrm{ appears in some } f \in F_L\right\}\\
    \label{eq:X2}
    X_N &=& X\setminus X_L.
\end{eqnarray}
The set $X_L$ contains the unknowns that appear in linear equations.
The set $X_N$ contains the unknowns that appear in equations of higher degree only.
We fix the following notation: $|F_L| = m_L,|F_N| = m_N,|X_L| = n_L,|X_N| = n_N$, which means that $m = m_L+m_N$ and $n = n_L + n_N$.

Now, the idea of our new elimination method is to eliminate all unknowns $X_N$ 
from non-linear equations $F_N$.
The non-linear equations $F_N$ do not depend on image measurements, and are the same for all instance of a given problem. Therefore, we can perform this elimination off-line, in the pre-processing step. This elimination may be computationally demanding. However, since we do this only once in the pre-processing step, it is not an issue during the solver run time. Next, we further eliminate $m_L$ unknowns from $X_L$ using $m_L$ linear equations from $F_L$. This is done  on-line but it is fast since solving a small linear system is easy. 

In more detail, our method performs the following steps.

\noindent {\bf Offline:}
\vspace{-0.2cm} 
\begin{enumerate}
\itemsep0em 
 \item Let $I = \langle F_{N} \rangle$ and consider the elimination ideal $I_{X_L} = I \cap \C[X_L]$.
 \item Compute the generators $G$ of $I_{X_L}$.  These contain unknowns from $X_L$ only, i.e.\ the unknowns appearing in the linear equations $F_L$.
 \end{enumerate}
 \vspace{-0.2cm} 
 {\bf Online:}
 \vspace{-0.2cm} 
 \begin{enumerate}
 \setcounter{enumi}{2}
 \itemsep0em 
 \item Rewrite the linear equations $F_L$ in the unknowns $X_L$ as~ $\M{M}X_L = 0$, where $\M{M}$ is a coefficient matrix and the vector $X_L$ contains all unknowns from $X_L$.  
 \item Compute a null space basis $\M{N}$ of $\M{M}$ and re-parametrize the unknowns $X_L = \M{N}\,Y$. If the rank of $\M{M}$ is $m_L$, i.e.\ the equations in $F_L$ are linearly independent, $Y$ would contain $k = n_L-m_L$ new unknowns. Note that if all input equations in $F$ were homogeneous, we could set one of the unknowns in $Y$ to 1 (assuming it is non-zero) and then $k = n_L-m_L-1$.
 \item Substitute $X_L = \M{N}\,Y$ into the generators $G$ of the elimination ideal $I_{X_L}$.
 \item Solve the new system of polynomial equations $G(Y)=0$ (e.g.\ using the \gb method and the precomputed elimination template for $G(Y)=0$ obtained by using the automatic generator~\cite{Kukelova-ECCV-2008}).
 \item Back-substitute to recover $X_{L} = \M{N}Y$.
 \item Extend partial solutions for $X_L$ to solutions for $X$.
\end{enumerate}

The main difference between our elimination strategy and the elimination strategies used before in minimal solvers (see Section~\ref{sec:CV_strategy}) is that the previous strategies substitute the parametrization $X_L = \M{N}\,Y$ directly into the input nonlinear equations $F_N$. This results in $m_N$ polynomial equations in $n_N + k$ unknowns $Y \cup X_N$.
On the other hand, the new method eliminates $n_N$ unknowns from the non-linear equations and creates a system $G(Y)=0$ in $k$ unknowns in the pre-processing step.
We will show on several important problems from computer vision
that solving the system $G(Y)=0$, instead of the system $F_N(Y \cup X_N)=0$, is more efficient.

Before presenting our new strategy on more complicated problems from computer vision, we illustrate the key ideas of our strategy on a simpler, but still representative, example. In 
Appendix~\ref{sec:homography}
we will show that the problem of estimating a 3D planar homography with unknown focal length, \ie the projection matrix with unknown focal length, from planar points leads to a system of polynomial equations with the same structure as in this illustrative example.
%
\subsection{Example} \label{subsec:Example}
\noindent Let us consider the following system of nine homogeneous polynomial equations in ten unknowns $X = \{h_1, h_2, h_3, h_4, h_5, h_6, h_7, h_8, h_9, w \}$. There are seven linear homogeneous equations in $h_1,\dots,h_9$, namely 
\begin{eqnarray}
F_L = \{f_j = \sum_{i=1}^{9} c_{ij} h_i = 0, \; j = 1,\dots,7; \; c_{ij} \in \mathbb{Q} \},
\label{eq:ex_FL}
\end{eqnarray}
and two $4^{th}$ order equations in~$\{h_1,h_2,h_4,h_5,h_7,h_8, w \}$ 
\begin{align}
\label{eq:ex_FN}
F_N = \{
w^2 h_1 h_2 + w^2 h_4 h_5 + h_7 h_8 = 0,  \qquad \qquad \quad \\ 
w^2h_1^2 + w^2h_4^2 + h_7^2 - w^2 h_2^2 - w^2 h_5^2 -h_8^2 =0 \nonumber\}. 
\end{align}
Using the notation from Section~\ref{sec:method} we have
\begin{eqnarray}
    X_L =  \{h_1,\dots,h_9\} \mbox{ and } X_N =  \{w\}.
\end{eqnarray}
We proceed as follows:
\begin{enumerate}
\itemsep0em 
 \item Create the elimination ideal 
 $$I_{w}\,\, = \,\, I\, \cap \,\mathbb{Q}[h_1,h_2,h_3,h_4,h_5,h_6,h_7,h_8,h_9].$$
 \item Compute the generator of the principal ideal $I_{w}$.
 This is a polynomial of degree four:
 \begin{align}
 \label{eq:ex_G}
 G = \{h_1h_2h_7^2+h_4h_5h_7^2-h_1^2h_7h_8+h_2^2h_7h_8 \nonumber \\ -h_4^2h_7h_8  +h_5^2h_7h_8-h_1h_2h_8^2-h_4h_5h_8^2\}
 \end{align}
 The polynomial in $G$ can be computed in the off-line pre-processing phase using the following code in the computer algebra system {\tt Macaulay2}~\cite{M2}:
\small
\begin{verbatim}
R = QQ[w,h1,h2,h3,h4,h5,h6,h7,h8,h9];
G = eliminate({w}, ideal(w^2*h1*h2 + 
w^2*h4*h5 + h7*h8, w^2*h1^2 + w^2*h4^2 + 
h7^2 - w^2*h2^2 - w^2*h5^2 - h8^2));
\end{verbatim}
\normalsize
  \item Rewrite seven linear equations form $F_L$~\eqref{eq:ex_FL} as $\M{M}\,h = 0$,  where $h = \left[ h_1,h_2,h_3,h_4,h_5,h_6,h_7,h_8,h_9\right]$ and $\M{M} = [c_{ij}]$ is \ntm{7}{9} coefficient matrix.
 \item Use a null space basis $\{n_1,n_2\}$ of $\M{M}$ to reparametrize the unknowns from $X_L$ with two unknowns as
 \begin{eqnarray}
 \label{eq:ex_param}
 h = y_1\,n_1 + y_2\,n_2.
 \end{eqnarray}
 Since the input equations are homogeneous, we set $y_2 = 1$ (assuming $y_2 \neq 0$).
 \item Substitute the new parametrization~\eqref{eq:ex_param} into the generator~\eqref{eq:ex_G}.
 \item Solve the resulting equation in one unknown $y_1$.
 \item Use the solutions for $y_1$ to recover solutions for $X_L$ using~\eqref{eq:ex_param}.
 \item Extend the solutions for $X_L$ to solutions for $X$ by substituting solutions to $F_N$.
\end{enumerate}
In this case, our elimination strategy generates one equation of degree four in one unknown.

On the other hand, the elimination strategy described in Section~\ref{sec:CV_strategy} generates two equations in two unknowns.
More precisely, the strategy from Section~\ref{sec:CV_strategy} would substitute parametrization~\eqref{eq:ex_param} directly into two equations from $F_N$~\eqref{eq:ex_FN}. 
This results in two equations in two unknowns $y_1$ and $w$. Solving this system of two equations in two unknowns in the on-line phase takes more time than solving a single quadratic equation.
\section{Applications}
\label{sec:applications}
\subsection{f+E+f relative pose problem}
\label{sec:fEf}
\noindent The first problem that we solve using our elimination strategy is that
 of estimating relative pose and the common unknown focal length of two cameras from six image point correspondences. This problem  is also known as the 6pt focal length problem, or the \fFf problem. The \fFf problem is a classical and popular problem in computer vision with many applications, \eg, in structure-from-motion~\cite{Snavely-IJCV-2008}.

The minimal \fFf problem has 15 solutions and it was first solved by Stew\`{e}nius \etal\cite{DBLP:journals/ivc/SteweniusNKS08} using the \gb method.
The 
solver of Stew\`{e}nius consists of three G-J eliminations of three matrices of size
\ntm{12}{33}, \ntm{16}{33} and  \ntm{18}{33} and the eigenvalue computation for a \ntm{15}{15} matrix.

More recently, two \gb solvers for the \fFf problem ware proposed in \cite{Byrod-ICCV-2007} and  \cite{Kukelova-ECCV-2008}.
The solver from~\cite{Byrod-ICCV-2007} performs SVD decomposition of a \ntm{34}{50}
matrix and it uses special techniques for improving the numerical stability of \gb solvers. 
The \gb solver generated by the automatic generator~\cite{Kukelova-ECCV-2008} performs G-J elimination of a \ntm{31}{46} matrix and is, to the best of our knowledge,
 the fastest and the most numerically stable solver for the \fFf problem.

All the state-of-the-art (SOTA) solvers exploit that the \ntm{3}{3} fundamental matrix $\mathtt{F} = \left[f_{ij}\right]_{i,j=1}^{3} \in\mathbb{R}^{3 \times 3}$  satisfies
\begin{eqnarray}
\M{E} = \M{K}^{\top}\M{F}\,\M{K} = \M{K}\,\M{F}\,\M{K}
\label{eq:KFK}
\end{eqnarray}
where $\M{K} = diag(f,f,1)$ is the diagonal  $3\times 3$ calibration matrix with the unknown focal length $f$ and $\M{E}$ is the $3 \times 3$ essential matrix~\cite{HZ-2003}. The essential matrix has rank 2 and satisfies the Demazure equations~\cite{Demazure88}  
\begin{eqnarray}
2\M{E}\,\M{E}^{\top}\M{E}-trace(\M{E}\,\M{E}^{\top})\M{E}=\M{0}.
\label{eq:trace}
\end{eqnarray}
(also known as the trace constraint).

In all 
SOTA
solvers~\cite{DBLP:journals/ivc/SteweniusNKS08,Byrod-ICCV-2007,Kukelova-ECCV-2008}, the linear equations from the epipolar constraints
\begin{eqnarray}
\V{x}_i^{\top}\M{F}\,\V{x}_i^{\prime} = 0
\label{eq:epipolar}
\end{eqnarray}
for six image point correspondences $\V{x}_i, \V{x}_i^{\prime}, \, i =1,\dots,6,$ in two views are first rewritten in a matrix form 
\begin{eqnarray}
\M{M}\,\M{f} = \V{0},
\end{eqnarray}
where $\M{M}$ is a \ntm{6}{9} coefficient matrix and $\M{f}$ is a vector of 9 
elements of the fundamental matrix $\M{F}$.
For six (generic) image correspondences in two views, the coefficient matrix $\M{M}$ has a three-dimensional null space.
Therefore, the fundamental matrix can be parametrized by two unknowns as
\begin{eqnarray}
\M{F} = x\,\M{F}_1 + y\,\M{F}_2 + \M{F}_3,
\label{eq:paramF}
\end{eqnarray}
where $\M{F}_1,\M{F}_2,\M{F}_3$ are matrices created from the three-dimensional null space of $\M{M}$ and $x$ and $y$ are new unknowns.

We use the parametrization~(\ref{eq:paramF}), the rank constraint for the fundamental matrix
\begin{eqnarray}
\det(\M{F})= 0,
\label{eq:detF}
\end{eqnarray}
and the trace constraint~(\ref{eq:trace})  for the essential matrix,
together with~(\ref{eq:KFK}) in the following form:
\begin{eqnarray}
2\,\M{F}\,\M{Q}\,\M{F}^{\top}\M{Q}\,\M{F}-trace(\M{F}\,\M{Q}\,\M{F}^{\top}\M{Q})\,\M{F}=\M{0},
\label{eq:traceF}
\end{eqnarray}
This  results in ten third- and fifth-order polynomial equations in
 three unknowns $x, y$ and  $w = 1/f^2$. In~(\ref{eq:traceF}) 
 we set $\M{Q} = \M{K}\M{K} = diag(f^2,f^2,1)$. 
 We note that the 
  trace constraint~(\ref{eq:traceF}) can be simplified by multiplying it with $1/w^2$. 
 
 The ten equations~(\ref{eq:detF}) and~(\ref{eq:traceF}) in three unknowns $x,y$ and $w$ were used as the input equations in all SOTA \gb solvers to  the \fFf problem~\cite{DBLP:journals/ivc/SteweniusNKS08,Byrod-ICCV-2007,Kukelova-ECCV-2008}.
 
 Note, that all  SOTA solvers followed the elimination method described in Section~\ref{sec:CV_strategy} and they differ only in the method used for solving the final non-linear system $F_N$.
 
Next, we present a new solver for the \fFf problem created using our elimination 
strategy in Section~\ref{sec:method}.
This strategy not only generates a more efficient solver, but it also reveals new
interesting  constraints  on  the  fundamental matrices of  cameras with unknown focal length.

\myparagraph{Elimination ideal formulation} 
For the \fFf problem we start with the ideal 
 $I \in \C\left[f_{11},f_{12},f_{13},f_{21},f_{22},f_{23},f_{31},f_{32},f_{33},f\right]$
generated by ten equations from the rank constraint~(\ref{eq:detF}) and the trace constraint~(\ref{eq:trace}) with the essential matrix~(\ref{eq:KFK}).

Since the epipolar constraint~(\ref{eq:epipolar}) gives us linear equations in $X_L = \left\{f_{11},f_{12},f_{13},f_{21},f_{22},f_{23},f_{31},f_{32},f_{33}\right\}$, we have $X_N = \{f\}$.
Hence the strategy presented in Section~\ref{sec:method} will first eliminate the unknown focal length $f$.

To compute the generators of the elimination ideal $I_f = I \cap  \C\left[f_{11},f_{12},f_{13},f_{21},f_{22},f_{23},f_{31},f_{32},f_{33}\right]$, i.e.\ the elements that do not contain the focal length $f$, 
we use the following {\tt Macaulay2}~\cite{M2} code:
{\small
\begin{verbatim}
R = QQ[f,f11,f12,f13,f21,f22,f23,f31,f32,f33];
F = matrix {{f11,f12,f13},{f21,f22,f23},
{f31,f32,f33}};
K = matrix {{f,0,0},{0,f,0},{0,0,1}};
E = K*F*K;
I = minors(1,2*E*transpose(E)*E
    -trace(E*transpose(E))*E)+ideal(det(E));
G = eliminate({f},saturate(I,ideal(f)))
dim G, degree G, mingens G
\end{verbatim}}
\normalsize
The output tells us that the variety of $G$ has  dimension $6$  and degree $15$,
and that $G$   is  the complete intersection of  two hypersurfaces in $\mathbb{P}^8$, cut out by the cubic 
\begin{equation}
\label{eq:cubic}
    {\rm  det}(\M{F})
\end{equation}
and the quintic
\begin{small}
\begin{equation}
 \begin{matrix}
f_{11} f_{13}^3 f_{31} +f_{13}^2 f_{21} f_{23} f_{31}+f_{11}f_{13} f_{23}^2 f_{31}
+f_{21}f_{23}^3 f_{31}\\
- f_{11} f_{13} f_{31}^3 - f_{21} f_{23}f_{31}^3  +  
 f_{12} f_{13}^3 f_{32} +f_{13}^2 f_{22} f_{23} f_{32} + \\
 f_{12} f_{13} f_{23}^2 f_{32} +f_{22} f_{23}^3 f_{32}-f_{12} f_{13} f_{31}^2 f_{32} - f_{12}^2 f_{13}^2 f_{33} \\
-f_{11} f_{13} f_{31} f_{32}^2- f_{21} f_{23} f_{31} f_{32}^2-f_{12}f_{13} f_{32}^3
{-}f_{22} f_{23} f_{32}^3\\
{-}f_{11}^2 f_{13}^2 f_{33} 
{-}f_{22} f_{23} f_{31}^2 f_{32} 
-2 f_{11} f_{13} f_{21}  f_{23} f_{33}-\\
2 f_{12} f_{13} f_{22} f_{23} f_{33} -
f_{21}^2 f_{23}^2 f_{33}   -  f_{22}^2 f_{23}^2 f_{33} +\\
   f_{11}^2 f_{31}^2 f_{33}  +f_{21}^2 f_{31}^2 f_{33}
  +2 f_{11}f_{12}f_{31} f_{32}f_{33} + \\
  2 f_{21} f_{22} f_{31} f_{32}f_{33}
  +f_{12}^2 f_{32}^2 f_{33} + f_{22}^2 f_{32}^2 f_{33}.
      \end{matrix}
\label{eq:quintic}
\end{equation}
\end{small}
The vanishing of ~(\ref{eq:cubic}) and~(\ref{eq:quintic}),
together with the equation for extracting the unknown focal length from the fundamental matrix~\cite{Newsam96} 
(see also Section~\ref{sec:focal_fEf})
completely describe the \fFf problem. Therefore we can formulate the following result.
\begin{result}
The zero set of~(\ref{eq:cubic}) and~(\ref{eq:quintic}) equals the space of all fundamental matrices $\M{F}$, i.e.\ the singular \ntm{3}{3} matrices, that can be decomposed into $\M{F} = \M{K}^{-1}\M{E}\,\M{K}^{-1}$,
where
$\M{K} = diag(f,f,1)$ for some non-zero $f \in \mathbb{C}$ and $\M{E}$ is an essential matrix.
By intersecting this variety with six hyperplanes given by the epipolar constraints~\eqref{eq:epipolar} for six image point correspondences, we obtain up to 15 real solutions for the fundamental matrix
(see Figure~\ref{fig:teaser}).
\end{result}

In our new efficient on-line solver for the \fFf problem, we first use the linear equations from the epipolar constraint~(\ref{eq:epipolar}) for six image point correspondences to parametrize the fundamental matrix $\M{F}$ with two new unknowns $x$ and $y$~(\ref{eq:paramF}). After substituting this parametrization into  the two generators~(\ref{eq:cubic}) and~(\ref{eq:quintic}) of $I_f$, we get two equations (of degree 3 and 5) in two unknowns $x$ and $y$. By solving these two equations we get up to 15 real solutions for the fundamental matrix $\M{F}$.

These two equations in two unknowns can be solved either using a Sylvester resultant~\cite{Cox-IVA-2015} or using the \gb method, which was used in all SOTA solvers for the \fFf problem. The \gb solver for these two equations, generated using the automatic generator~\cite{Kukelova-ECCV-2008}, performs G-J elimination of a \ntm{21}{36} matrix. This matrix contains almost $3\times$ less nonzero elements than the matrix from the smallest \ntm{31}{46} SOTA solver~\cite{Kukelova-ECCV-2008} that was also generated using the automatic generator, however for the original formulation with ten equations in three unknowns. The sparsity patterns of both these solvers are shown in
Section~\ref{sec:sparsity}.

\myparagraph {Experiments}
Since the new solver for the \fFf problem is algebraically equivalent to the SOTA solvers, we have evaluated the new \fFf solver on synthetic noise free data only.

\begin{figure}[t]
\centering
\begin{tabular}{cc}
\includegraphics[width=0.52\linewidth]{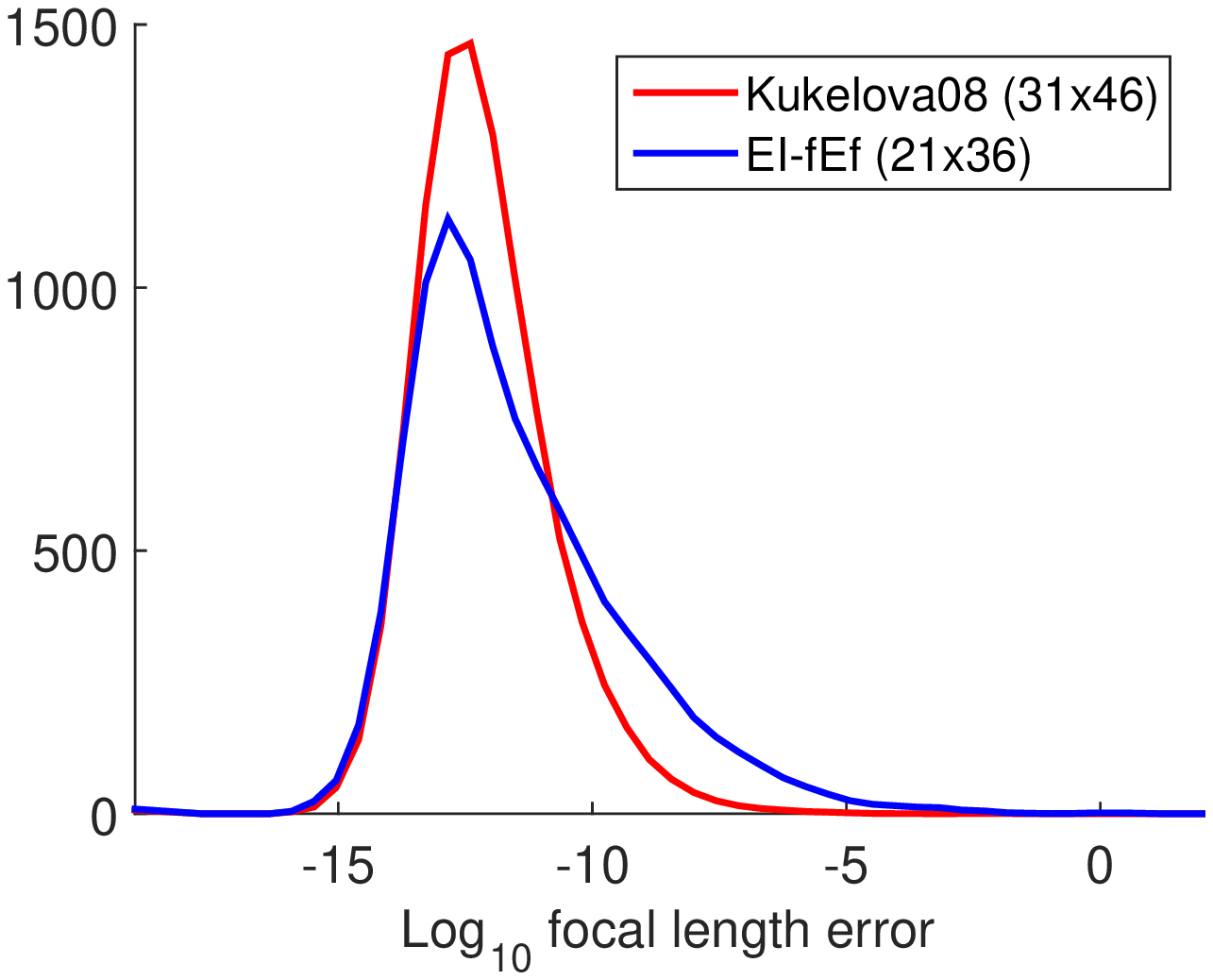}
\hspace{-0.5cm} &
\hspace{-0.5cm} 
\includegraphics[width=0.52\linewidth]{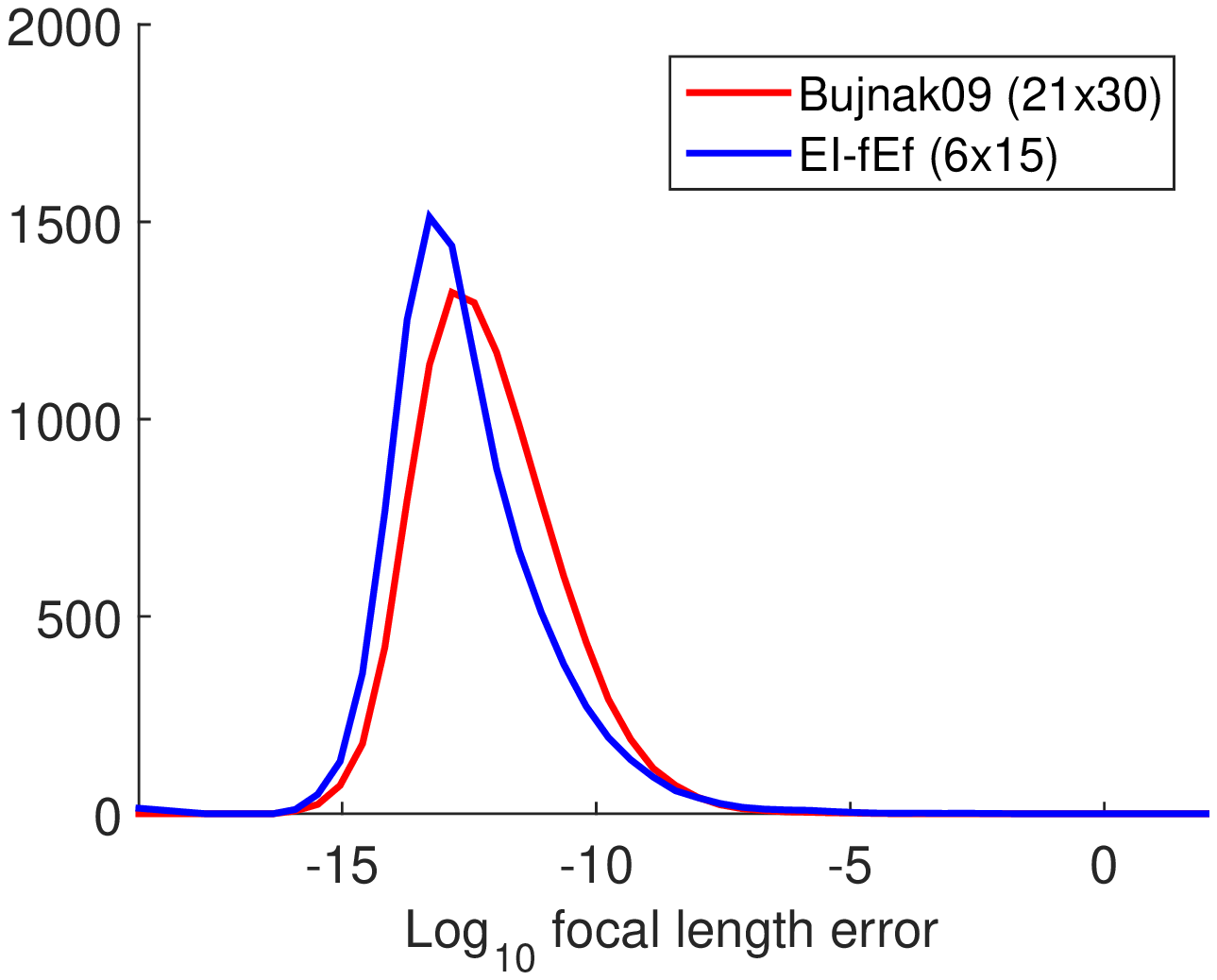}
\vspace{-0.2cm} 
\tabularnewline
\vspace{-0.0cm} 
{\scriptsize{(a)}} &
{\scriptsize{(b)}} 
\end{tabular}
\caption{{\bf Numerical stability:}
$\textrm{Log}_{10}$ of the relative error of the focal length for the (a) \fFf problem; EI-fEf solver (blue), Kukelova08~\cite{Kukelova-ECCV-2008} (red). (b) \Ef problem; EI-Ef solver (blue), Bujnak09~\cite{Bujnak-ICCV-2009} (red).}
\label{fig:6pt}
\end{figure}

We studied the behavior of the new \fFf solver (EI-fEf) based on the new elimination strategy presented in Section~\ref{sec:method}  on noise-free data to check its numerical stability. We compared it to the results of the SOTA \gb solver Kukelova08~\cite{Kukelova-ECCV-2008}. 
In this experiment, we generated 10000 synthetic scenes with 3D points distributed at random in a $\left[-10,10\right]^3$ cube. Each 3D point was projected by two cameras with random but feasible orientation and position and with random focal length $f_{gt} \in \left[0.5, 5\right]$.
Figure~\ref{fig:6pt}(a) shows $\log_{10}$ of the relative error
of the focal length $f$ obtained 
by selecting the real root closest to the ground truth value $f_{gt}$. In the case of the new EI-fEf solver, the focal length $f$ was extracted from the computed $\M{F}$ using the formula presented in
Section~\ref{sec:focal_fEf}.
However, any method for extracting the focal length from $\M{F}$, e.g. the SVD-based method~\cite{Newsam96}, can be used here.

The new EI-fFf solver (blue) is slightly less stable than Kukelova08 (red). 
However, both solvers provide very stable results without larger errors and in the presence of noise and in real applications their performance is in fact equivalent.
Moreover, the new solver is smaller and more efficient.

\subsection{E+f 6pt relative pose problem} \label{subsec:E+f}
\noindent
The second problem that we solve using the new elimination strategy is the problem of estimating relative pose of one calibrated and one up to focal length calibrated camera from six image point correspondences, \ie  the \Ef problem.

The minimal \Ef problem was first solved by Bujnak~\etal~\cite{Bujnak-ICCV-2009} using 
Gr\"obner bases 
and the polynomial eigenvalue method. Their solver performs G-J eliminations on a 
\ntm{21}{30} matrix and the eigenvalue computation for a \ntm{9}{9} matrix.

For the \Ef problem the first camera is calibrated up to an unknown focal
length and the second camera is fully calibrated. Therefore, the relationship between the  essential and the fundamental matrix has the form
\begin{eqnarray}
\M{E} = \M{F}\,\M{K},
\label{eq:Ef}
\end{eqnarray}
where  $\M{K}=diag(f,f,1)$ is a diagonal calibration matrix of the first camera, containing the unknown focal length $f$. By substituting this relationship into the trace constraint for the essential matrix~(\ref{eq:trace}),  and setting
$\M{Q} = \M{K}\,\M{K} $,
we obtain
\begin{eqnarray}
2\,\M{F}\,\M{Q}\,\M{F}^{\top}\M{F}-trace(\M{F}\,\M{Q}\,\M{F}^{\top})\M{F}=\M{0}.
\label{eq:traceEf}
\end{eqnarray}
 
The SOTA solver~\cite{Bujnak-ICCV-2009}  uses the elimination strategy from Section~\ref{sec:CV_strategy} and starts by rewriting the epipolar constraint~(\ref{eq:epipolar}) as
$\M{M}\,\V{f} = \V{0}$.
Then, the fundamental matrix is parametrized by two unknowns $x$ and $y$ as in the \fFf case (Eq.~(\ref{eq:paramF})). With this formulation, the rank constraint~(\ref{eq:detF}) and the trace constraint~(\ref{eq:traceEf}) result in ten third and fourth order polynomial equations in three unknowns $x,y$ and $w=1/f^2$. These ten equations are solved in~\cite{Bujnak-ICCV-2009} by using the automatic generator of \gb solvers~\cite{Kukelova-ECCV-2008}.

Next we present a new solution to the \Ef problem that uses the new elimination strategy from Section~\ref{sec:method}.

\myparagraph{Elimination ideal formulation}
We start with the ideal 
 $I \in \C\left[f_{11},f_{12},f_{13},f_{21},f_{22},f_{23},\right.$
 $\left. f_{31},f_{32},f_{33},f\right]$
generated by ten equations from the rank constraint~(\ref{eq:detF}) and the trace constraint~(\ref{eq:trace}), with the essential matrix~(\ref{eq:Ef}).
As for the \fFf problem, the epipolar constraint~(\ref{eq:epipolar}) gives linear equations in $X_1 = \left\{f_{11},f_{12},f_{13},f_{21},f_{22},f_{23},f_{13},f_{23},f_{33}\right\}$.  Therefore we again will eliminate only the unknown focal length $f$.

To compute the generators of the elimination ideal $I_f = I \cap  \C\left[f_{11},f_{12},f_{13},f_{21},f_{22},f_{23},f_{31},f_{32},f_{33}\right]$, i.e.\ the generators that do not contain $f$, we can use a similar \texttt{Macaulay2} code as for the \fFf problem, just by replacing  line  {\tt E = K*F*K} with line {\tt E = F*K}. 

For the \Ef  problem, the variety of $G$ has dimension $6$ and degree $9$ in $\mathbb{P}^8$  and is defined by one cubic and three quartics 
(see Appendix~\ref{sec:generators}). 


In the online solver, the epipolar constraint~\eqref{eq:epipolar} for six image point correspondences is used to parametrize the fundamental matrix $\M{F}$ with two new unknowns $x$ and $y$~(\ref{eq:paramF}). 
This parametrization, applied to the four generators of the elimination ideal $I_f$, gives four equations of degree three and four in two unknowns. We solve these four equations in two unknowns using the \gb method~\cite{Kukelova-ECCV-2008}. The \gb solver, generated using the automatic generator~\cite{Kukelova-ECCV-2008}, performs G-J elimination of a \ntm{6}{15} matrix. This matrix is much smaller than the elimination template matrix from the SOTA solver~\cite{Bujnak-ICCV-2009}, which has the size \ntm{21}{30}.

\myparagraph {Experiments}
We studied the behavior of the new \Ef elimination ideal based solver (EI-Ef) on noise-free data and compared it to the results of the SOTA \gb solver Bujnak09~\cite{Bujnak-ICCV-2009}. 

We generated 10000 synthetic scenes with 3D points distributed at random in a $\left[-10,10\right]^3$ cube. Each 3D point was projected by two cameras with random but feasible orientation and position. The focal length of the first camera was randomly drawn from the interval $f_{gt} \in \left[0.5, 5\right]$ and the focal length of the second camera was set to 1, i.e. the second camera was considered as calibrated. Figure~\ref{fig:6pt}(b) shows $\log_{10}$ of the relative error of the focal length $f$ obtained by selecting the real root closest to the ground truth value $f_{gt}$. For the new EI-Ef solver, the focal length $f$ was extracted from the computed $\M{F}$ using the formula presented in 
Appendix~\ref{sec:focal_Ef}.

The new EI-Ef solver (blue) is not only smaller but also slightly more stable than Bujnak09~\cite{Bujnak-ICCV-2009} (red). Both solvers provide very stable results without larger errors.

\subsection{E+f+k 7pt relative pose problem}
\label{sec:Efk}
\noindent
The last problem that we will formulate and solve using the new elimination strategy presented in Section~\ref{sec:method} is the problem of estimating the epipolar geometry of one calibrated camera and one camera with unknown focal length and unknown radial distortion, i.e.\  uncalibrated camera with radial distortion. We denote this problem by \Efk.

A popular model for radial distortion is the one-parameter division model~\cite{Fitzgibbon-CVPR-2001}. This is an undistortion mo- del that can handle even quite pronounced radial distortions:
\begin{equation}
	\label{eq:undist_eq}
	\V{x}_{u_{i}}(\lambda)                     =
	\left[{x}_{d_{i}},{y}_{d_{i}},1+\lambda({x}_{d_{i}}^2+{y}_{d_{i}}^2)\right]^{\top}.
\end{equation}
In this model $\V{x}_{d_{i}} = \left[x_{d_{i}},{y}_{d_{i}},1\right]^{\top}$ are the
homogeneous coordinates of the measured (and radially distorted) image points and $\lambda\in\mathbb{R}$ is the distortion parameter.
This model was used in the first 7pt minimal solution to the \Efk problem 
presented in~\cite{Kuang-CVPR-14}.

For the \Efk problem, the epipolar constraint reads as
\begin{equation} 
\label{eq:epipolarEfk} \quad
\V{x}_{u_{i}}^\top
\mathtt{F}\,\V{x}_{u_{i}}^{\prime}\left( \lambda \right) = 0,
\qquad i=1,\ldots,7,
\end{equation} 
where $\mathbf{x}_{u_{i}}, \mathbf{x}_{u_{i}}^{\prime}\left( \lambda \right) \in\mathbb{R}^{3}$ 
are the homogeneous coordinates of corresponding ideally projected image points, \ie, points not corrupted by radial distortion~\cite{HZ-2003}. 
Note that for the right camera we do not know the camera calibration parameters and we measure distorted image points. Therefore, to use these distorted image points in the epipolar constraint, we first need to undistort them using the model~(\ref{eq:undist_eq}). 

The epipolar constraint~(\ref{eq:epipolarEfk}) together with the trace~(\ref{eq:traceEf}) and the rank constraint~(\ref{eq:detF}) form a quite complicated system of polynomial equations. Note that all equations in this system are non-linar and therefore the method from Section~\ref{sec:CV_strategy} cannot be directly applied.

In the SOTA solver~\cite{Kuang-CVPR-14}, this system is first simplified by manually eliminating some unknowns. First, the authors set $f_{33} = 1$, which implies that their solver does not work for motions where $f_{33} = 0$. Next, they use the epipolar constraint~(\ref{eq:epipolarEfk}) for six image point correspondences to eliminate six unknowns $f_{11}, f_{12}, f_{21}, f_{22}, f_{31}, f_{32}$, which appear linearly in the epipolar constraint, from the equations. Then, the remaining equation from the epipolar constraint for the seventh image point correspondence, together  with the trace~(\ref{eq:traceEf}) and the rank constraint~(\ref{eq:detF}), form a system of 11  (one quadratic, four $5^{th}$ and six $6^{th}$ degree) equations in four unknowns $f_{13}, f_{23}, \lambda, w= 1/f^2$. Then, the equations are again manually simplified.
They generate the elimination template by multiplying 11 input equations by a set of monomials such that  the maximum degree of the monomials in the resulting equations is 8. The resulting elimination
template has size \ntm{200}{231}. The authors of this solver observed that by using automatic strategies from~\cite{Kukelova-ECCV-2008} or by further reducing the size of the elimination template, the numerical stability of their solver deteriorates. To improve the numerical stability of the final \gb solver, the authors further choose 40 monomials instead of necessary 19 for basis selection.

It can be seen that the \Efk problem requires a very careful manual manipulation of the input equations to get a numerically stable solver. However, still, the final solver is quite large and not really useful in real applications. 

Here we will show that using the new elimination strategy presented in Section~\ref{sec:method} we can solve this problem efficiently without the need for any special manipulation and treatment of input equations. Moreover, the final solver obtained using this new method is much more efficient and numerically stable than the SOTA solver~\cite{Kuang-CVPR-14}.

\myparagraph{Elimination ideal formulation} 
Unfortunately, for the \Efk problem the epipolar constraint~(\ref{eq:epipolarEfk}) does not give us linear equations. Therefore, we can't directly apply the method presented in Section~\ref{sec:method}.
However, in this case we can easily linearize the equations from the epipolar constraint~(\ref{eq:epipolarEfk}). 

The epipolar constraint~(\ref{eq:epipolarEfk})  
contains monomials $(f_{11}, f_{12}, f_{13}, f_{21}, f_{22}, f_{23}, f_{31}, f_{32}, f_{33},
f_{13} \lambda,  f_{23} \lambda, f_{33} \lambda)$. To linearize the equations~(\ref{eq:epipolarEfk}) we set 
\begin{eqnarray}
\label{eq:linearization1}
y_{13} =f_{13} \lambda, \\
\label{eq:linearization2}
y_{23} = f_{23} \lambda, \\
\label{eq:linearization3}
y_{33} = f_{33} \lambda.
\end{eqnarray}
Now the equations from ~(\ref{eq:epipolarEfk}) can be seen as linear homogeneous equations in the 12 unknowns $f_{11},f_{12},f_{13},f_{21},f_{22},f_{23},f_{31},f_{32},f_{33}, y_{13},y_{23},y_{33}$.

Another view on this linearization is that the distorted image points are lifted to 4D space and the fundamental matrix $\M{F}$ is enriched by one column to 
\begin{eqnarray}
\label{eq:Flift}
\hat{\M{F}}= 
\begin{pmatrix} 
      \, \M{F} | \V{f}_3 \lambda
     \end{pmatrix} = 
\begin{pmatrix}
     \,f_{11} &  f_{12}  & f_{13} & y_{13}  \\
     \,f_{21} & f_{22} & f_{23}  &  y_{23}\\
     \,f_{31}  & f_{32} & f_{33} &  y_{33}
     \end{pmatrix},
\end{eqnarray}
where $\V{f}_3$ is the $3^{rd}$ column of $\M{F}$. The \ntm{3}{4} fundamental matrix $\hat{\M{F}}$~(\ref{eq:Flift}) was introduced in~\cite{Brito-BMVC12} and is known  as the one-sided radial distortion matrix.
With this matrix, the epipolar constraint~(\ref{eq:epipolarEfk}) can be written as
\begin{eqnarray}
\label{eq:epipolar_lift}
\V{x}_{u_{i}}^\top
\mathtt{F}\,\V{x}_{u_{i}}^{\prime}\left( \lambda \right) = 
\V{x}_{u_{i}}^\top
\hat{\M{F}}\left[{x}^{\prime}_{d_{i}},{y}^{\prime}_{d_{i}},1,{x}_{d_{i}}^{\prime \,2}+{y}_{d_{i}}^{\prime\, 2}\right]^{\top} = 0.
\end{eqnarray}

For the \Efk problem, our method starts with the ideal 
 $I \! \in \! \C\left[f_{11},f_{12},f_{13}, f_{21},f_{22},f_{23},f_{31},f_{32},f_{33}, y_{13}, \right.$ $\left. y_{23}, y_{33}, \lambda, f\right]$
generated by 13 equations, i.e.\ three equations from the constraints~(\ref{eq:linearization1})-(\ref{eq:linearization3}), the rank constraint~(\ref{eq:detF}) and the nine equations from the trace constraint~(\ref{eq:trace}), with the essential matrix of the form~(\ref{eq:Ef}).
These 13 equations form the set $F_N$ from ~\eqref{eq:F1} in our elimination strategy.

In this case, the ``lifted'' epipolar constraint~(\ref{eq:epipolar_lift}) gives us linear equations in 12 elements of the $3 \times 4$ radial distortion fundamental matrix $\hat{\M{F}}$~(\ref{eq:Flift}), i.e.\ linear equations in $X_L = \left\{f_{11},f_{12},f_{13},f_{21},f_{22},f_{23},f_{31},f_{32},f_{33},y_{13},y_{23},y_{33}\right\}$. 
Therefore, for the \Efk problem, we use the new elimination strategy to eliminate two unknowns, the focal length $f$ and the radial distortion parameter $\lambda$, \ie $X_N = \{f,\lambda\}$.

To compute the generators of the elimination ideal $I_{f,\lambda} = I \cap  \C\left[f_{11},f_{12},f_{13},f_{21},f_{22},f_{23},f_{31},f_{32},f_{33},\right.$ $\left.y_{13},y_{23},y_{33}\right]$, i.e.\ the generators that do not contain 
$f$ and $\lambda$,
we can use a similar \texttt{Macaulay2} code as for the \Ef problem. 
We only need to replace the first line with 
\small
\begin{verbatim}
R = QQ[f,k,f11,f12,f13,f21,f22,f23,f31,f32,
f33,y13,y23,y33];
\end{verbatim}
\normalsize
and add one additional line at the end
\small
\begin{verbatim}
Gu = eliminate({k}, G +
ideal(y13-f13*k,f23-f23*k,f33-f33*k,))
codim Gu, degree Gu, mingens Gu
\end{verbatim}
\normalsize
For the \Efk  problem the variety  of
{\tt Gu} has dimension $7$ and degree $19$ in $\mathbb{P}^{11}$. 
In addition to the three quadrics of the form 
$f_{i3}\,y_{j3} - f_{j3}\,y_{i3}$, the ideal generators for
{\tt Gu} are two cubics and nine quartics, i.e. altogether 14 polynomials.
Although this system of 14 polynomial equations looks quite complex it is much easier to solve than the original system with $\lambda$ and $f$ that was used in the SOTA~\cite{Kuang-CVPR-14}.

The ``lifted'' epipolar constraint~\eqref{eq:epipolar_lift} for seven general image point correspondences can be rewritten as $\M{M}\hat{\V{f}}=\V{0}$, where $\M{M}$ is \ntm{7}{12} coefficient matrix and $\hat{\V{f}}$ is a \ntm{12}{1} vector containing the elements of the one-sided distortion fundamental matrix  $\hat{\M{F}}$.
This means that the one-sided distortion fundamental matrix $\hat{\M{F}}$ can be parametrized by four new unknowns $x_1, x_2, x_3$ and $x_4$, using the 5-dimensional null space of~$\M{M}$, as 
\begin{eqnarray}
\hat{\M{F}} = x_1\hat{\M{F}}_1 + x_2\hat{\M{F}}_2+  x_3\hat{\M{F}}_3+  x_4\hat{\M{F}}_4+  \hat{\M{F}}_5.
\label{eq:F_lift_lin}
\end{eqnarray}

Substituting (\ref{eq:F_lift_lin}) into the 14 generators of the elimination ideal $I_{f,\lambda}$
gives $14$ equations in four unknowns. 
We solve these equations using the \gb method~\cite{Kukelova-ECCV-2008}.
The \gb solver, generated using the automatic generator~\cite{Kukelova-ECCV-2008}, performs G-J elimination of a \ntm{51}{70} matrix.
This matrix is  much smaller than the elimination template matrix from the SOTA solver~\cite{Kuang-CVPR-14}, which has the size \ntm{200}{231}.
Moreover, the new solver doesn't require an application of the methods for improving numerical stability of the \gb solver that were used in~\cite{Kuang-CVPR-14}.

After solving 14 equations in four unknowns $x_1,x_2,x_3,x_4$, we reconstruct solutions for $\M{F}$ using~\eqref{eq:F_lift_lin}, solutions for $\lambda$ using~\eqref{eq:linearization1}, and solutions for $f$ using the formula presented in 
Appendix~\ref{sec:focal_Ef}.

\begin{figure}[t]
\centering
\begin{tabular}{cc}
\includegraphics[width=0.52\linewidth]{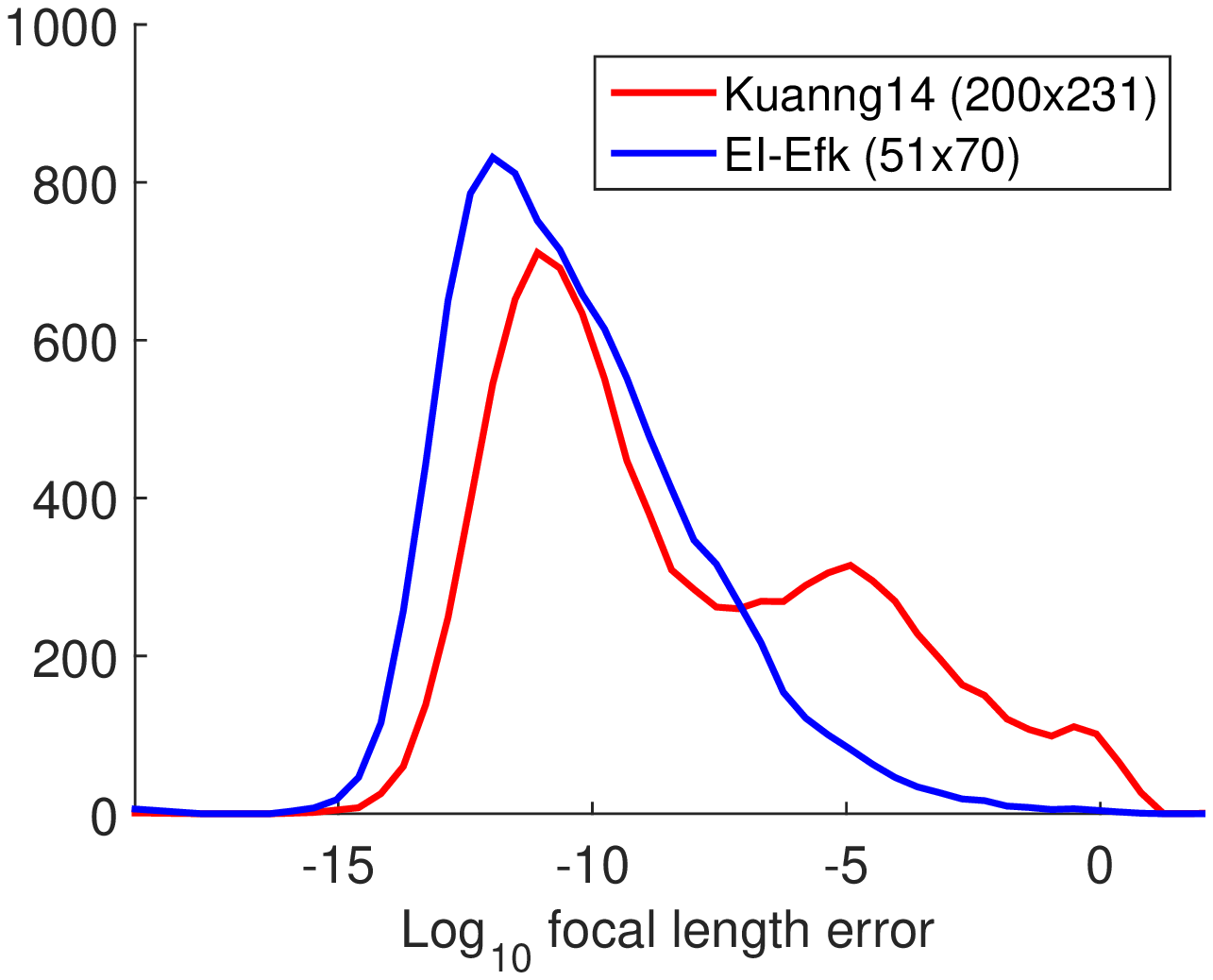}
\hspace{-0.5cm} &
\hspace{-0.5cm} 
\includegraphics[width=0.52\linewidth]{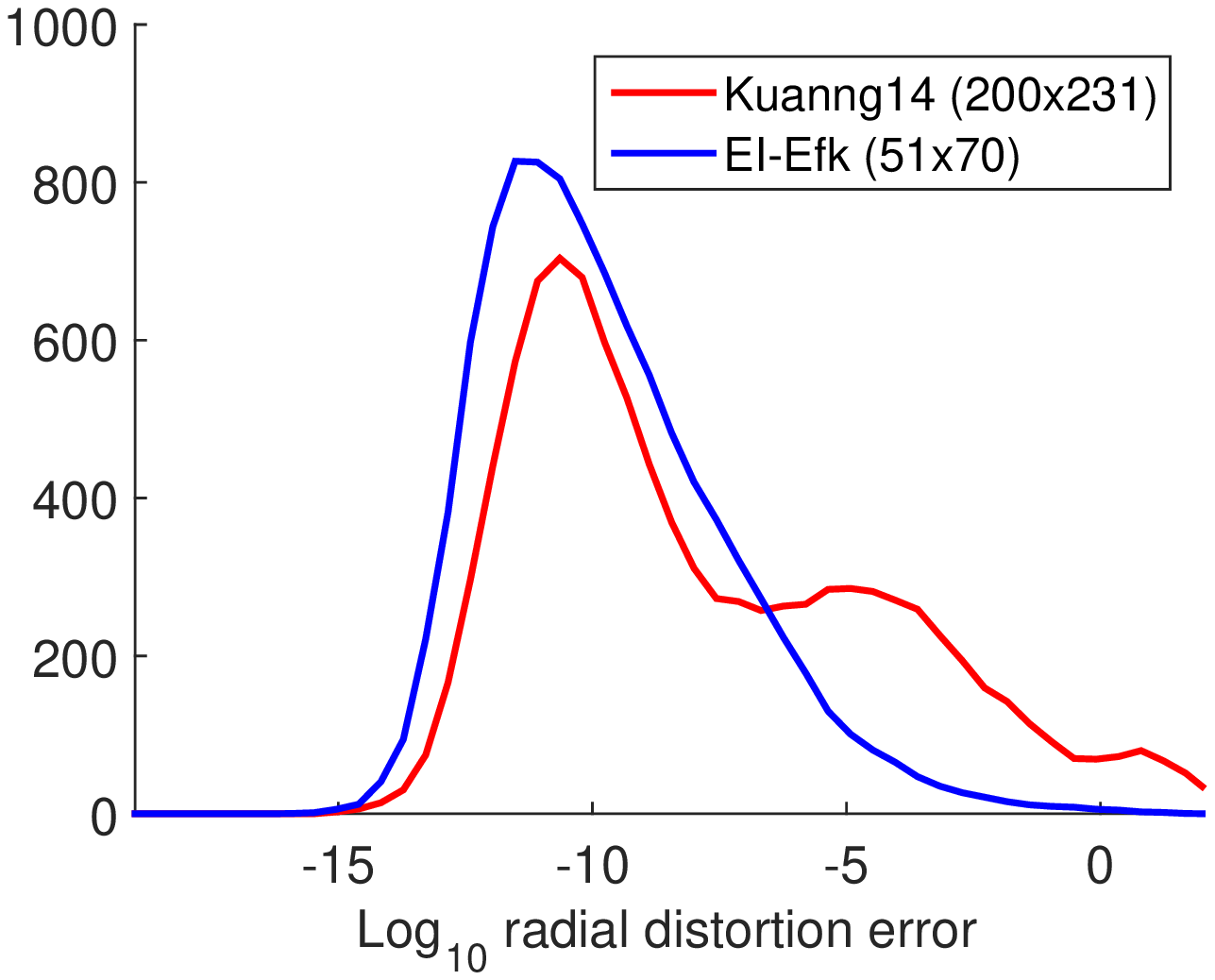}
\vspace{-0.2cm} 
\tabularnewline
\vspace{-0.0cm} 
{\scriptsize{(a)}} &
{\scriptsize{(b)}} 
\end{tabular}
\caption{{\bf Numerical stability \Efk problem :} (a) $\textrm{Log}_{10}$ of the relative error of the focal length  (b) $\textrm{Log}_{10}$ of the relative error of the radial distortion; EI-Efk solver (blue), Kuang14~\cite{Kuang-CVPR-14}(red).}
\label{fig:7pt}
\end{figure}
\myparagraph{Experiments}
We first studied the numerical stability of the new \Efk solver (EI-Efk) on noise-free data  and compared it to the results of the SOTA \gb solver Kuang14~\cite{Kuang-CVPR-14}. In this experiment, we generated 10000 synthetic scenes in the same way as in the \Ef experiment and image points in the first camera were corrupted by radial distortion following the one-parameter division model. The radial distortion parameter $\lambda_{gt}$ was drawn at random from
the interval $[-0.7, 0]$. 
Figure~\ref{fig:7pt}(a) shows $\log_{10}$ of the relative error
of the focal length $f$ obtained by selecting the real root closest to the ground truth value $f_{gt}$. 
Figure~\ref{fig:7pt}(b) shows $\log_{10}$ of the relative error
of the radial distortion $\lambda$ obtained by selecting the real root closest to the ground truth value $\lambda_{gt}$. 
In this case the new EI-Efk solver (blue) is not only significantly smaller but also significantly more stable than the SOTA \gb solver Kuang14~\cite{Kuang-CVPR-14} (red). 
What is really important for real applications is that the new EI-Efk solver provides very stable results without larger errors, while for the SOTA solver Kuang14~\cite{Kuang-CVPR-14} we observe many failures.

\begin{figure}[t]
\setlength{\belowcaptionskip}{-100pt}
\begin{minipage}[c]{0.25\textwidth}
\includegraphics[width=1.1\linewidth]{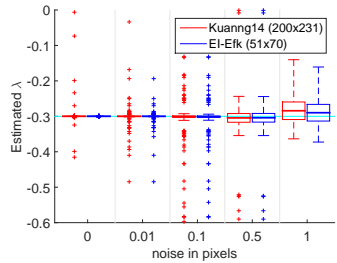}
\hspace{-0.5cm}
\end{minipage}\hfill
 \begin{minipage}[c]{0.2\textwidth}
\caption{%
\small Comparison of the new EI-Efk solver (blue) with the SOTA  Kuang14~\cite{Kuang-CVPR-14} solver (red). Boxplots of estimated $\lambda$'s for different noise levels and 	$\lambda_{gt} = -0.3$
\label{fig:noise_Efk}
}
\end{minipage}
\vspace{-1cm}
\end{figure}

Next, Figure~\ref{fig:noise_Efk} shows the results of experiments with noise simulation for the \Efk problem. We show the estimated radial distortion parameters for the ground truth radial distortion $\lambda_{gt} = -0.3$ and 200 runs for each noise level. We compared our new \Efk solver with the SOTA Kuang solver~\cite{Kuang-CVPR-14}. Figure~\ref{fig:noise_Efk} shows resuts by {\sc Matlab} {\tt boxplot}. In the presence of noise, our new EI-Efk solver (blue) gives similar or even better estimates than the SOTA solver Kuang14~\cite{Kuang-CVPR-14} for which we observed more failures (crosses). 
\subsection{Computational complexity}

\noindent Here, we show a comparison of the computational efficiency of the new elimination-based solvers (EI-fEf, EI-Ef, EI-Efk) and the SOTA solvers~\cite{Kukelova-ECCV-2008,Bujnak-ICCV-2009,Kuang-CVPR-14}. Since we do not have comparable implementations of the SOTA solvers, we compare the sizes of the G-J eliminations (QR decompositions) performed by these solver. G-J elimination is one of the most time consuming steps for all solvers. The comparison of sizes is reported in the Tab.~\ref{tab:sizes}. The last row of this table displays the ratio of the number of non-zero elements of the template matrices of SOTA solvers ($nz_S$) and the number of non-zero elements of the template matrices of our new elimination-based solvers ($nz_{EI}$).
\begin{table}[htbp]
\begin{center}
\begin{tabular}{|l|l|l|l|}
\hline 
 & \fFf
 & \Ef
 & \Efk \tabularnewline
\hline 
\hline
SOTA & \ntm{31}{46}~\cite{Kukelova-ECCV-2008} & \ntm{21}{30}~\cite{Bujnak-ICCV-2009} & \ntm{200}{231}~\cite{Kuang-CVPR-14}\tabularnewline
\hline 
EI (new) & \ntm{21}{36} & \ntm{~~6}{15} & \ntm{~~51}{70}\tabularnewline
\hline
\hline 
$nz_{S} / nz_{EI}$ & ~~~~3 & ~~~~5.2 & ~~~~2.8 \tabularnewline
\hline
\end{tabular}
\end{center}
\caption{New EI solvers are much smaller than SOTA solvers. Lines SOTA vs EI~(new) show that clever solvers eliminate much smaller matrices. They also manipulate much fewer numbers. See ratios of non-zero numbers $nz_S/nz_{EI}$ in the SOTA ($nz_S$) vs new ($nz_{EI}$) solvers.}
\label{tab:sizes}
\end{table}

\section{Conclusion}
\noindent We have presented a new insight into minimal solver construction based on elimination theory. By eliminating separately linear and non-linear equations and combining that later, we were able to generate much smaller solvers than before, see Tab.~\ref{tab:sizes}. We also generated an interesting new constraint, Eq.~\eqref{eq:quintic}, on partially calibrated camera pairs. Our method was first motivated by the idea of exploiting linear equations~\eqref{eq:F1} of our systems but we also demonstrated that it can produce efficient solvers~(Sec.~\ref{sec:Efk}) by linearizing fully non-linear situations.
%

%
%
\section{Appendix}\label{sec:appendix}
\noindent This appendix includes (1) additional details on the Elimination Theorem 
(in Sec.~\ref{sec:ElimThorem}), (2) derivation of the constraints on the projection for planar scenes by cameras with unknown focal length 
(in Sec.~\ref{sec:homography}), (3) details of focal length extraction 
(in Sec.~\ref{sec:focal}), (4) detailed presentation of the generators of $E+f$ problem 
(in Sec.~\ref{sec:generators}), and (5) results on the solvers' sparsity 
(in Sec.~\ref{sec:sparsity}).  
\subsection{Details on the elimination theorem}
\label{sec:ElimThorem}
\noindent Here we provide additional details for
\begin{theorem}[Elimination theorem~\cite{Cox-IVA-2015}]\label{thm:ElimTheorem}
Let $I \subseteq \C[x_1, \ldots , x_n]$ be an ideal and let G be a \gb of I with respect to the lexicographic monomial order where $x_1 > x_2 > \cdots > x_n$. Then, for every $0\leq l \leq n$, the set $G_l = G \cap\, \C[x_{l+1},\ldots,x_n]$ is a \gb of the l-th elimination ideal $I_l = I \cap \C[x_1, \ldots, x_l]$ .
\end{theorem}
See~\cite{Cox-IVA-2015} for a full account of the theory.

The {\em ring} $\C[x_1, \ldots , x_n]$ stands for all polynomials in $n$ unknowns $x_1, \ldots , x_n$ with complex coefficients. In computer vision applications, however, coefficients of polynomial systems are always real (in fact, rational) numbers and our systems consist of a finite number $s$ of polynomial equations $f_i(x_1, \ldots , x_n)=0$, $i=1,\ldots,s$.

The {\em ideal} $I = \{\sum_{i=1}^s h_i f_i \,|\, h_i, \ldots\ h_s \in \C[x_1, \ldots , x_n] \}$ generated by $s$ polynomials (generators) $f_i$ is the set of all polynomial 
linear combinations of the polynomials $f_i$. Here the multipliers $h_i$ 
are polynomials. All elements in the ideal $I$ evaluate to zero (are satisfied) 
at the solutions to the equations $f_i(x_1, \ldots , x_n)=0$.

The {\em \gb} $G = \{g_1, \ldots, g_m \}$ of an ideal $I$ is  a particularly convenient set of the generators of $I$, which can be used to find solutions to the original system $f_i$ in an easy way. For instance, for linear (polynomial) equations, a \gb of the ideal generated by the linear polynomials is obtained by Gaussian elimination. After Gaussian elimination, equations appear in a triangular form allowing one to solve for one unknown after another. This pattern carries on in a similar way to (some) \gbs of general polynomial systems and thus it makes \gbs a convenient tool for solving general polynomial systems.

Algorithmic construction of \gbs relies on an ordering of monomials to specify in which order to deal with monomials of a polynomial. {\em Lexicographic monomial order} (LEX) is a particularly convenient order, which can be used to produce \gbs that are in the triangular form. LEX orders monomials as words in a dictionary. An important parameter of a LEX order (i.e.\ ordering of words) is the order of the unknowns (i.e.\ ordering of letters). For instance, monomial $xy^2z = xyyz > xyzz = xyz^2$ when $x>y>z$ (i.e. $xyyz$ is before $xyzz$ in a standard dictionary). However, when $x<y<z$, then $xy^2z = xyyz < xyzz = xyz^2$. We see that there are $n!$ possible LEX orders when dealing with $n$ unknowns.  

The set $G_l = G \cap \C[x_{l+1},\ldots,x_n]$ contains all the polynomials in \gb $G$ that contain only unknowns $x_{l+1},\ldots,x_n$. For instance, if $G$ is a \gb in the triangular form, then $G_l = \{g_m(x_n), g_{m-1}(x_{n-1},x_n),\ldots,g_{m-l}(x_1,\ldots,x_{l+1})\}$ contains polynomials in one, two, \ldots, $l$ unknowns. 

The polynomials $G_l$ generate the {\em elimination ideal} $I_l = I \cap \C[x_{l+1},\ldots,n]$, containing all polynomials from $I$ that use the unknowns $x_{l+1},\ldots,x_n$ only. Hence, for each of $n!$ orderings, we get $n$ elimination ideals $I_l$.
\subsection{3D planar homograpy with unknown focal length}
\label{sec:homography}

\noindent We assume that a planar object (say, simply a plane) is observed by an unknown camera with the projection matrix~\cite{HZ-2003} 
\begin{equation}
\M{P} = \M{K}[\M{R}\,|\, \V{t}],    
\label{eq:P}
\end{equation}
where $\M{K} = diag(f,f,1)$ is the calibration matrix with the unknown focal length $f$, $\M{R} = [r_{ij}]_{ij=1}^3 \in SO(3)$ is the unknown rotation, and $\V{t} = [t_1,t_2,t_3]^{\top} \in \mathbb{R}^3$  the unknown translation.

Without loss of generality, we assume that the plane is defined by $z = 0$, \ie all 3D points with homogeneous coordinates $\V{X}_i = [x_i,y_i,z_i,1]^{\top}$ have the $3^{rd}$ coordinate $z_i = 0$.
Then, the image points $\V{u}_i = [u_i,v_i,1]^{\top}$ and the corresponding 3D points $\V{X}_i=\left[x_i,y_i,0,1\right]^{\top}$ are related by
\begin{equation}
	\label{eq:homography_eq}
	\alpha_i\, \V{u}_{i} = \M{H}\,\hat{\V{X}}_{i},
\end{equation}
where $\alpha_i$ are unknown scalars, $\hat{\V{X}}_{i} = [x_i,y_i,1]$,  and $\M{H} = [h_{ij}]_{ij=1}^{3} \in \mathbb{R}^{3 \times 3}$
is a homography matrix that has the form
\begin{eqnarray}
\M{H} = 
\mat{c}{\V{p}_1 \; \V{p}_2 \; \V{p}_4} = 
\mat{rrr}{
     f\, r_{11} &  f\, r_{12}  & t_{1}  \\
     f\, r_{21} &  f\, r_{22} &  t_{2}\\
     r_{31}  & r_{32} & t_{3}}
     \label{eq:H}
\end{eqnarray}
where $\V{p}_j$ is the $j^{th}$ column of the projection matrix $\M{P}~\eqref{eq:P}$.

Next, from the projection equation~\eqref{eq:homography_eq}, we eliminate the scalar values $\alpha_i$. This can be done by multiplying~\eqref{eq:homography_eq} by  the skew symmetric matrix
$\xx{\V{u}}$~\cite{HZ-2003}
to get
\begin{small}
\begin{equation}
 \label{eq:homography_eq_skew}
	\left[ 
		\begin{array}{rrr}
			0                    & -1 & v_i  \\ 
			1   & 0                   & -u_i\\  
			\!\!\! -v_i & u_i  & 0                   \\ 
		\end{array}
	\right] \!
	\left[\begin{array}{ccc}
\!		h_{11} & h_{12} & h_{13} \\ 
\!		h_{21} & h_{22} & h_{23} \\ 
\!		h_{31} & h_{32} & h_{33} 
		\end{array}
	\right] \!
	\left[ 
		\begin{array}{c}
			x_i \\ 
			y_i \\  
			1
		\end{array}
	\right]=\M{0} \,\,
\end{equation}
\end{small}
%
%
The matrix equation~\eqref{eq:homography_eq_skew} contains three polynomial
equations, two of which are linearly independent. This means that we need at least $3.5$ 2D $\leftrightarrow$ 3D point correspondences to estimate the unknown homography $\M{H}$, because $\M{H}$ has 7 degrees of freedom: three parameters for the rotation, three parameters for the translation and also the focal length.

For the $3.5$ point correspondences, matrix equation~\eqref{eq:homography_eq_skew} results in  seven linearly independent linear homogeneous   equations   in   nine  elements of the homography matrix $\M{H}$.

Moreover, we have here two additional polynomial constraints on elements of $\M{H}$.
For the first two columns of the rotation matrix $\M{R}$, there holds
\begin{eqnarray}
r_{11}r_{12} + r_{21}r_{22} + r_{31}r_{32} = 0\\
r_{11}^2 + r_{21}^2 + r_{31}^2 -  r_{12}^2 - r_{22}^2 - r_{32}^2  = 0 
\end{eqnarray}
This means that the elements of the first two columns of the homography matrix $\M{H} = [h_{ij}]_{ij=1}^{3}$~\eqref{eq:H} satisfy
\begin{eqnarray}
w^2\,h_{11}\,h_{12} + w^2\,h_{21}h_{22} +  h_{31}h_{32} = 0 \\
w^2\,h_{11}^2 + w^2\,h_{21}^2 + h_{31}^2 -  w^2\,h_{12}^2 - w^2\,h_{22}^2 - h_{32}^2  = 0
\end{eqnarray}
where $w = 1/f$.

Hence, estimating 3D planar homography with unknown focal length results in seven linear homogeneous equations and two non-linear homogeneous equations in $X = \{h_{11}, h_{12}, h_{13}, h_{21}, h_{22}, h_{23}, h_{31}, h_{32}, h_{33}, w \}$. This system of nine homogeneous equations has the same form as that presented in Section~\ref{subsec:Example}. Therefore this system can be efficiently solved using the new elimination  strategy presented in Section \ref{sec:method}. This strategy results in solving one fourth-degree equation in one unknown (see Section~\ref{subsec:Example}).
\subsection{Extraction of the focal length}
\label{sec:focal}
\noindent In this section we present formulas for extracting the focal length from a given fundamental matrix $\M{F}$ for two cases
\begin{enumerate}
    \item $\M{E} = \M{F}\,\M{K}$
    \item $\M{E} = \M{K}\,\M{F}\,\M{K}$ 
\end{enumerate}
where  $\M{K}=diag(f,f,1)$ is a diagonal calibration matrix.
Unlike most of the existing formulas and methods for extracting the focal length from the fundamental matrix $\M{F}$, the presented formulas contain directly elements of the fundamental matrix. They don't require an SVD decomposition of the fundamental matrix or computation of the epipoles.
\subsubsection{\Ef problem}
\label{sec:focal_Ef}
\noindent Here we will assume that the principal points~\cite{HZ-2003} are at the origin (which can be always achieved by shifting the known principal points) and use the recent result~\cite[Lemma 5.1]{KileelKPS16} which we restate in our notation:

\begin{lemma}
  Let $\M{F}$ be a fundamental matrix of the form that satisfies $\M{E} = \M{F}\,\M{K}$.  Then there are exactly two pairs of essential matrix and focal length $(\M{X} = \M{E},f)$ and  $(\M{X} = \textup{diag}(-1,-1,1)\M{E}, \, -f)$.  The positive $f$ is recovered from $\M{F} = [f_{ij}]_{1 \leq i, j \leq 3}$ by the following formula  
\begin{tiny}
\begin{equation*}
\label{eq:focal}
f^2=
    \frac{ f_{23}f_{31}^2+f_{23}f_{32}^{2}-2f_{21}f_{31}f_{33}-2f_{22}f_{32}f_{33}-f_{23}f_{33}^{2}}
    {2f_{11}f_{13}f_{21}+2f_{12}f_{13}f_{22}-f_{23}(f_{11}^{2}-f_{12}^{2}+f_{13}^{2}+f_{21}^{2}+f_{22}^{2}+f_{23}^2}
\end{equation*}
\end{tiny}
\end{lemma}
\subsubsection{\fFf problem}
\label{sec:focal_fEf}
\noindent To derive formulas for the extraction of $f$ from $\M{F}$ computed from images with the same unknown focal length, we follow methods developed in~\cite{KileelKPS16}. In this case, the result is the following formula for $f^2$, namely:
\begin{tiny}
$-f_{13}^2f_{32}f_{33}-f_{23}^2f_{32}f_{33}+f_{12}f_{13}f_{33}^2+f_{22}f_{23}f_{33}^2 $
\end{tiny}
quantity divided by
\begin{tiny}
$f_{11}f_{13}f_{31}f_{32}+f_{21}f_{23}f_{31}f_{32}+f_{12}f_{13}f_{32}^2+f_{22}f_{23}f_{32}^2-f_{11}f_{12}f_{31}f_{33}-f_{21}f_{22}f_{31}f_{33}-f_{12}^2f_{32}f_{33}-f_{22}^2f_{32}f_{33})$
\end{tiny},
which can be obtained by the following {\tt Macaulay2} code
{\small
\begin{verbatim}
R = QQ[f,f11,f12,f13,f21,f22,f23,f31,f32,f33]
F = matrix{{f11,f12,f13},{f21,f22,f23},
           {f31,f32,f33}};
K = matrix{{f, 0, 0}, {0, f, 0}, {0, 0, 1}};
E = K*F*K;
G = ideal(det(E))+minors(1,2*E*transpose(E)*E
          -trace(E*transpose(E))*E);
Gs = saturate(G,ideal(f));
gse = flatten entries mingens gb Gs;
cofs = g->coefficients(g,Variables=>{f});
cofsg = apply(gse,cofs);
cofsg_2
\end{verbatim}}

%
%
\subsection{The elimination ideal for the \Ef problem}
\label{sec:generators}
\noindent We consider the \Ef problem from Section~\ref{subsec:E+f}, \ie the problem of estimating epipolar geometry of one calibrated and one up to focal length calibrated camera.
Here, in this case
\begin{eqnarray}
\M{E} = \M{F}\,\M{K},
\label{eq:Ef_app}
\end{eqnarray}
where  $\M{K}=diag(f,f,1)$ is a diagonal calibration matrix for the first camera, containing the unknown focal length $f$.
Here, $\M{F}$ is the \ntm{3}{3} fundamental matrix and $\M{E}$ 
is the \ntm{3}{3} essential matrix~\cite{HZ-2003}

For the \Ef problem, we have the ideal 
 $I \subset \C\left[f_{11},f_{12},f_{13},f_{21},f_{22},f_{23},
f_{31},f_{32},f_{33},f\right]$
generated by ten equations, one cubic from the rank constraint
\begin{eqnarray}
\det(\M{F}) = 0,
\end{eqnarray}
and nine polynomials from the trace constraint
\begin{eqnarray}
2\,\M{F}\,\M{Q}\,\M{F}^{\top}\M{F}-trace(\M{F}\,\M{Q}\,\M{F}^{\top})\M{F}=\M{0},
\label{eq:traceEf_app}
\end{eqnarray}
 where $\M{Q} = \M{K}\,\M{K} $. 


For this problem, the new elimination strategy from Section~\ref{sec:method} leads to computing the generators of the elimination ideal $I_f = I \cap  \C\left[f_{11},f_{12},f_{13},f_{21},f_{22},f_{23},f_{31},f_{32},f_{33}\right]$, i.e.\ the generators that do not contain $f$.
To compute these generators we can use the following {\tt Macaulay2}~\cite{M2} code:
{\small
\begin{verbatim}
R = QQ[f,f11,f12,f13,f21,f22,f23,f31,f32,f33];
F = matrix {{f11,f12,f13},{f21,f22,f23},
{f31,f32,f33}};
K = matrix {{f,0,0},{0,f,0},{0,0,1}};
E = F*K;
I = minors(1,2*E*transpose(E)*E
    -trace(E*transpose(E))*E)+ideal(det(E));
G = eliminate({f},saturate(I,ideal(f)))
dim G, degree G, mingens G
\end{verbatim}}

For the \Ef  problem, the variety $G$ has dimension $6$ and degree $9$ in $\mathbb{P}^8$  and is defined by one cubic and three quartics.
It can be verified that these four polynomials correspond to the four maximal minors of the  \ntm{3}{4} matrix:
\begin{equation}
\label{eq:threebyfour}
 \begin{pmatrix}
     \,f_{11} &  f_{12}  & f_{13} & \,\,f_{21} f_{31}+f_{22} f_{32} + f_{23} f_{33} \\
     \,f_{21} & f_{22} & f_{23}  &  -f_{11} f_{31}-f_{12} f_{32}-f_{13} f_{33} \\
     \,f_{31}  & f_{32} & f_{33} & 0
     \end{pmatrix}.
\end{equation}     


\subsection{Sparsity patterns of solvers}
\label{sec:sparsity}
\noindent Here, we show a comparison of the sparsity patterns of our new elimination-based solvers (EI-fEf, EI-Ef, EI-Efk) and of the SOTA solvers~\cite{Kukelova-ECCV-2008,Bujnak-ICCV-2009,Kuang-CVPR-14}. 

\begin{figure}[t]
\centering
\begin{tabular}{cc}
\includegraphics[width=0.52\linewidth]{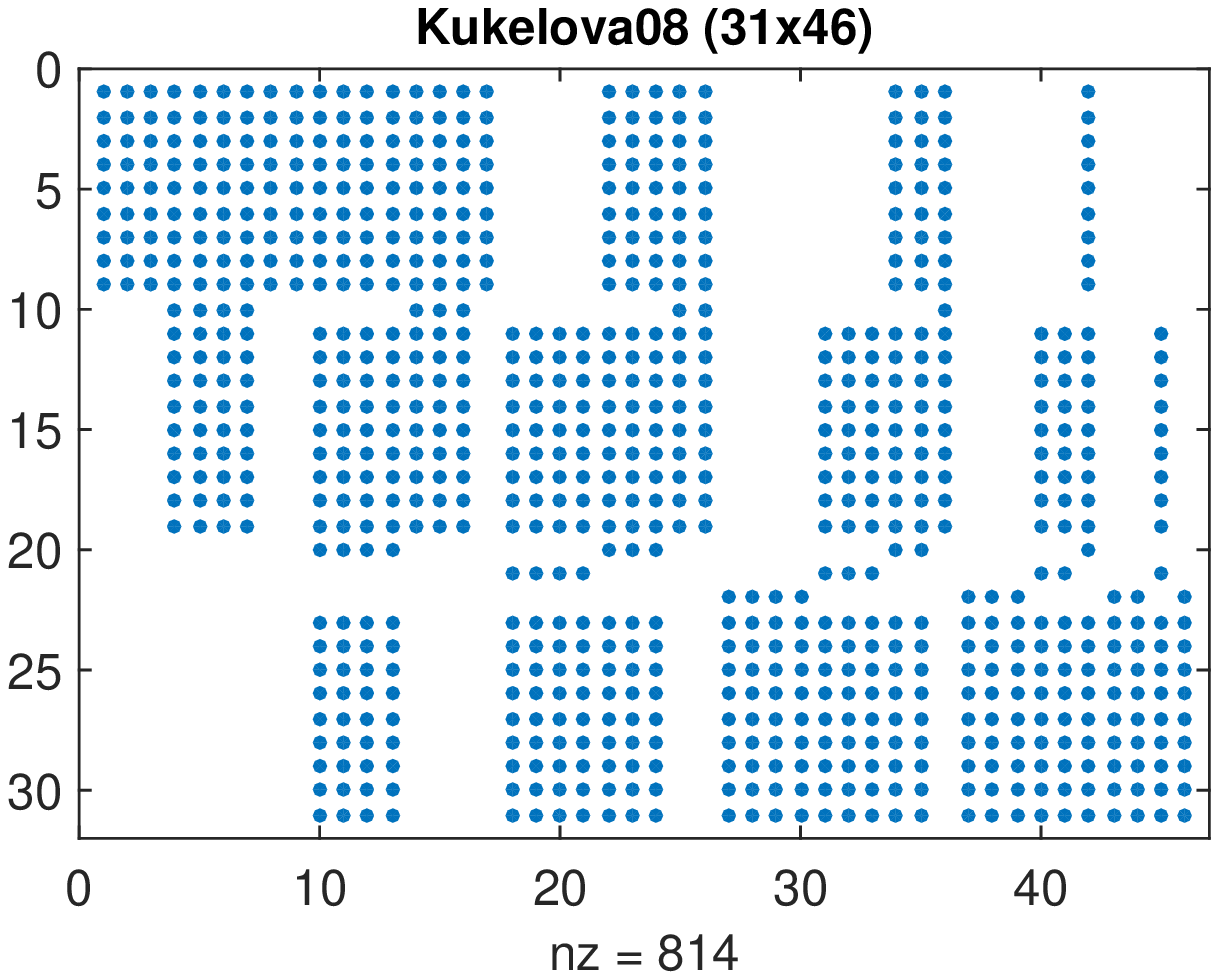}
\hspace{-0.3cm} &
\hspace{-0.3cm} 
\includegraphics[width=0.52\linewidth]{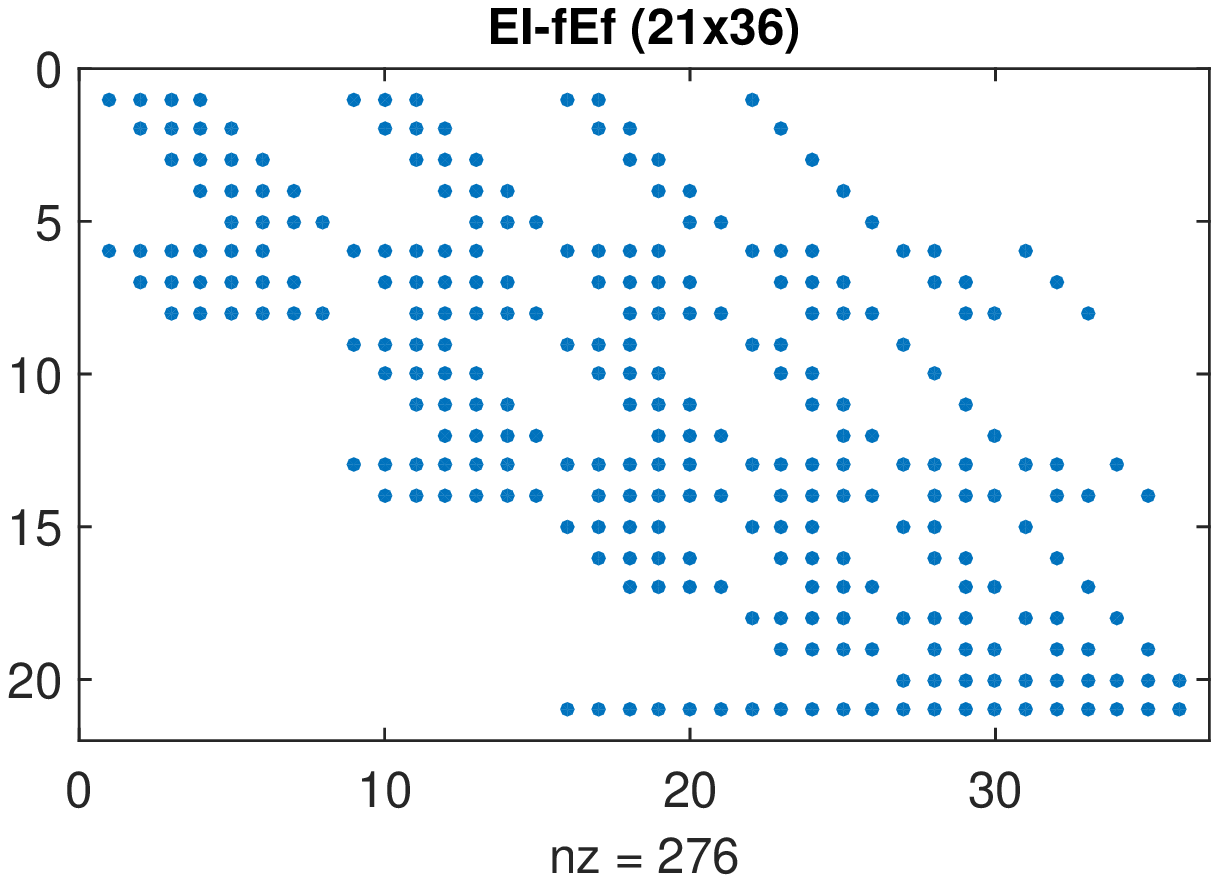}
\vspace{-0.2cm} 
\tabularnewline
\vspace{-0.0cm} 
{\scriptsize{(a)}} &
{\scriptsize{(b)}} 
\end{tabular}
\caption{{\bf Sparsity  patterns for the solvers to the \fFf problem:} (a)  state-of-the-art \ntm{31}{46} Kukelova08~\cite{Kukelova-ECCV-2008} solver (b) the new \ntm{21}{36} EI-fEf solver.}
\label{fig:fEf_sparsity}
\end{figure}

Figure~\ref{fig:fEf_sparsity} shows the sparsity patterns of the (a) state-of-the-art (SOTA) \ntm{31}{46} Kukelova08~\cite{Kukelova-ECCV-2008}  solver for the \fFf problem and (b) the new \ntm{21}{36} EI-fEf solver for this problem.
In this case the new EI-fEf solver is not only smaller but also sparser.
 The ratio of the number of non-zero elements of the \ntm{31}{46} template matrix of the SOTA solver Kukelova08~\cite{Kukelova-ECCV-2008} ($nz_S$) and the number of non-zero elements of the \ntm{21}{36} matrix of the EI-fEf solver ($nz_{EI}$) is 3.

\begin{figure}[t]
\centering
\begin{tabular}{cc}
\includegraphics[width=0.52\linewidth]{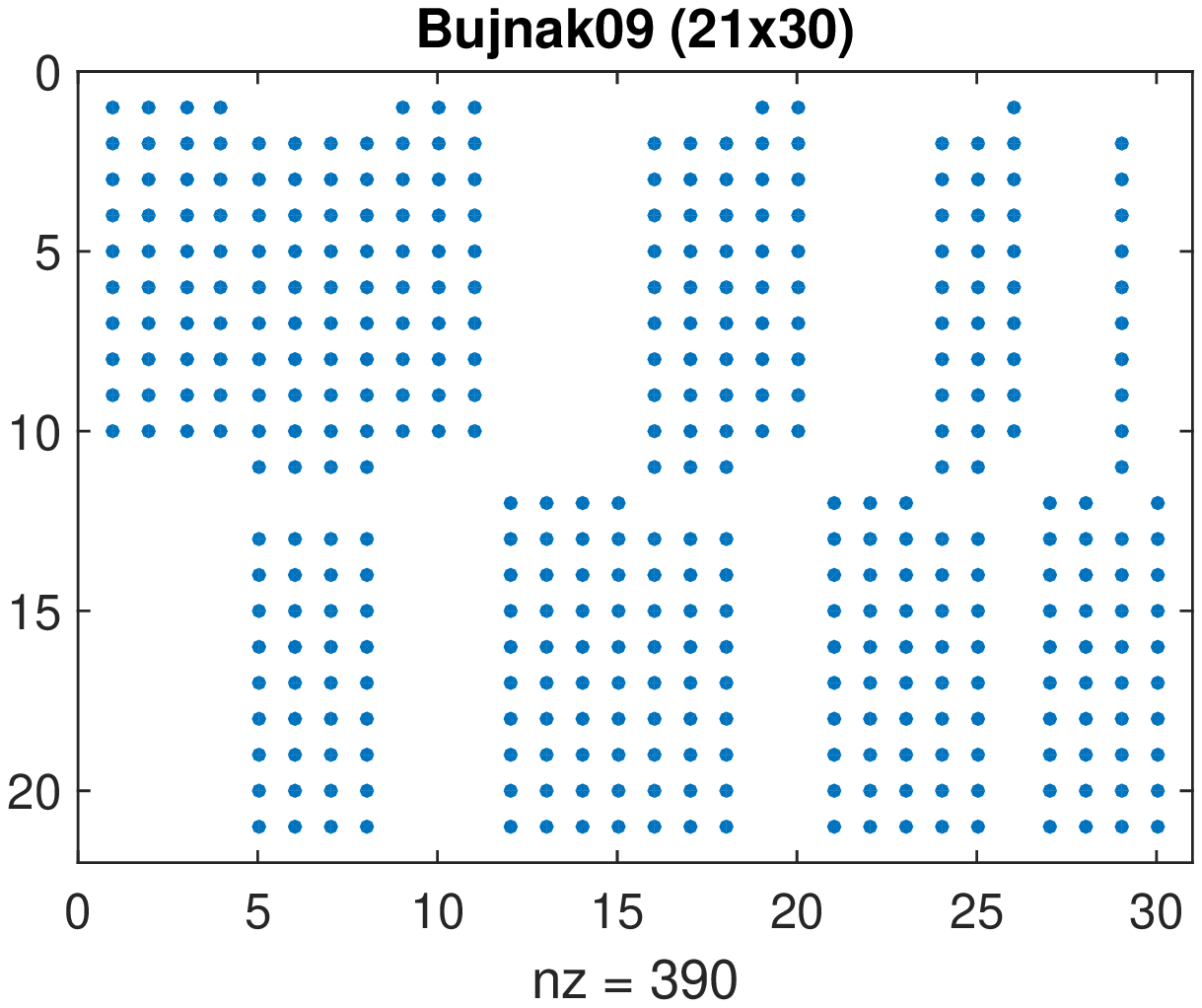}
\hspace{-0.3cm} &
\hspace{-0.3cm} 
\includegraphics[width=0.52\linewidth]{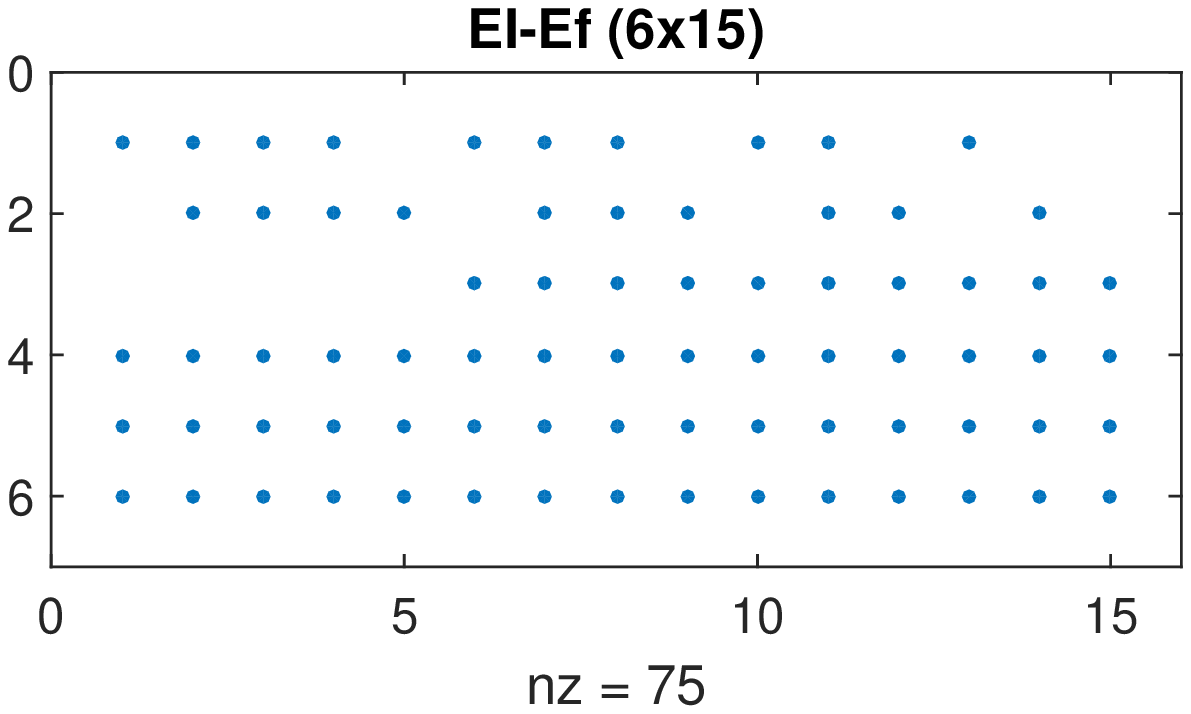}
\vspace{-0.2cm} 
\tabularnewline
\vspace{-0.0cm} 
{\scriptsize{(a)}} &
{\scriptsize{(b)}} 
\end{tabular}
\caption{{\bf Sparsity  patterns for the solvers to the \Ef problem:}  (a) state-of-the-art \ntm{21}{30} Bujnak09~\cite{Bujnak-ICCV-2009} solver (b) the new \ntm{6}{9} EI-Ef solver.}
\label{fig:Ef_sparsity}
\end{figure}

Figure~\ref{fig:Ef_sparsity} shows the sparsity patterns of the (a) SOTA \ntm{21}{30} Bujnak09~\cite{Bujnak-ICCV-2009} solver and (b) the new \ntm{6}{15} EI-Ef solver for the \Ef problem.
Here the ratio of the number of non-zero elements of the template matrix of the SOTA solver~\cite{Bujnak-ICCV-2009}  
 and the number of non-zero elements of the template matrix of our new   EI-Ef solver is 5.2.

\begin{figure}[t]
\centering
\begin{tabular}{cc}
\includegraphics[width=0.52\linewidth]{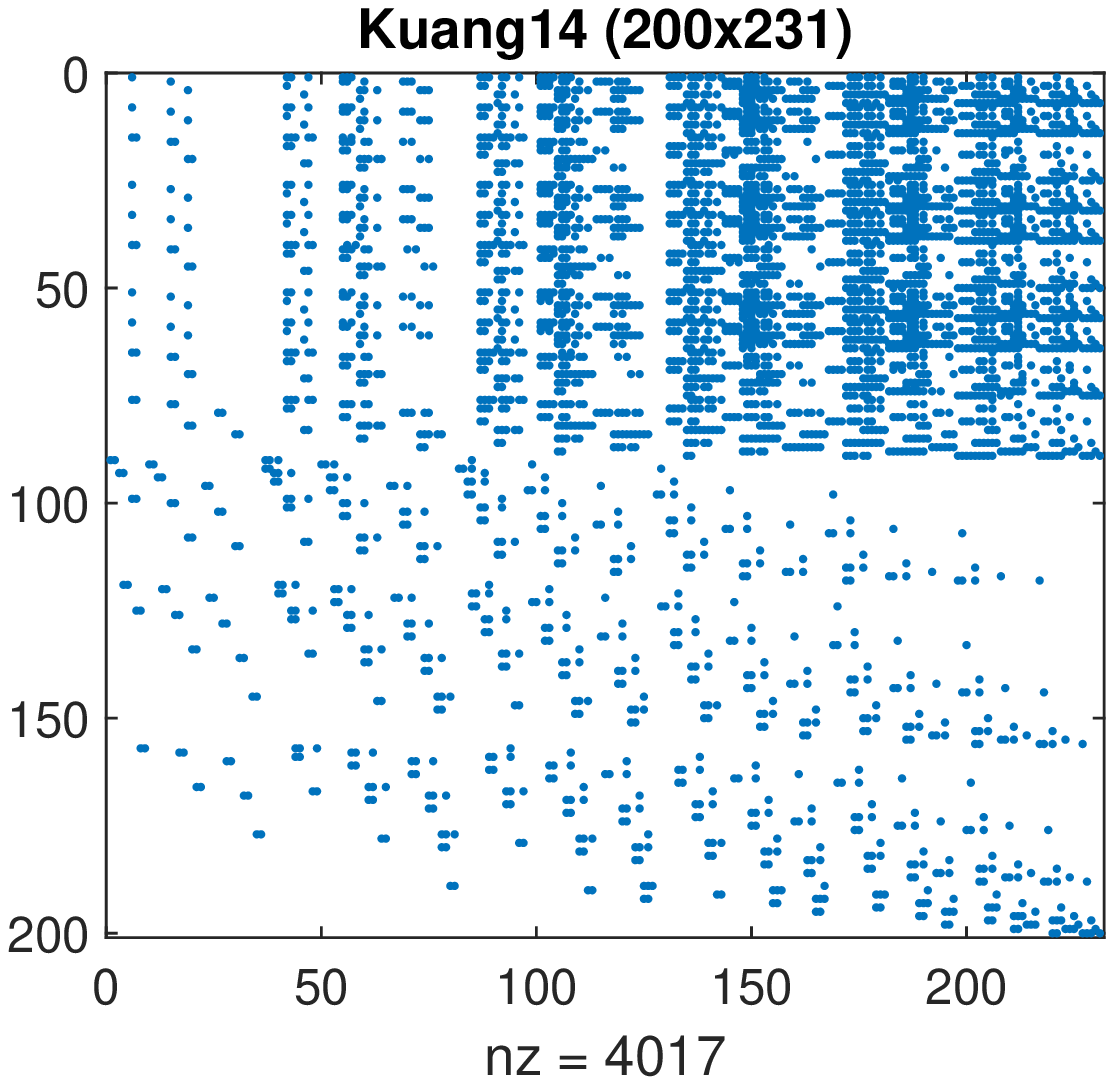}
\hspace{-0.3cm} &
\hspace{-0.3cm} 
\includegraphics[width=0.52\linewidth]{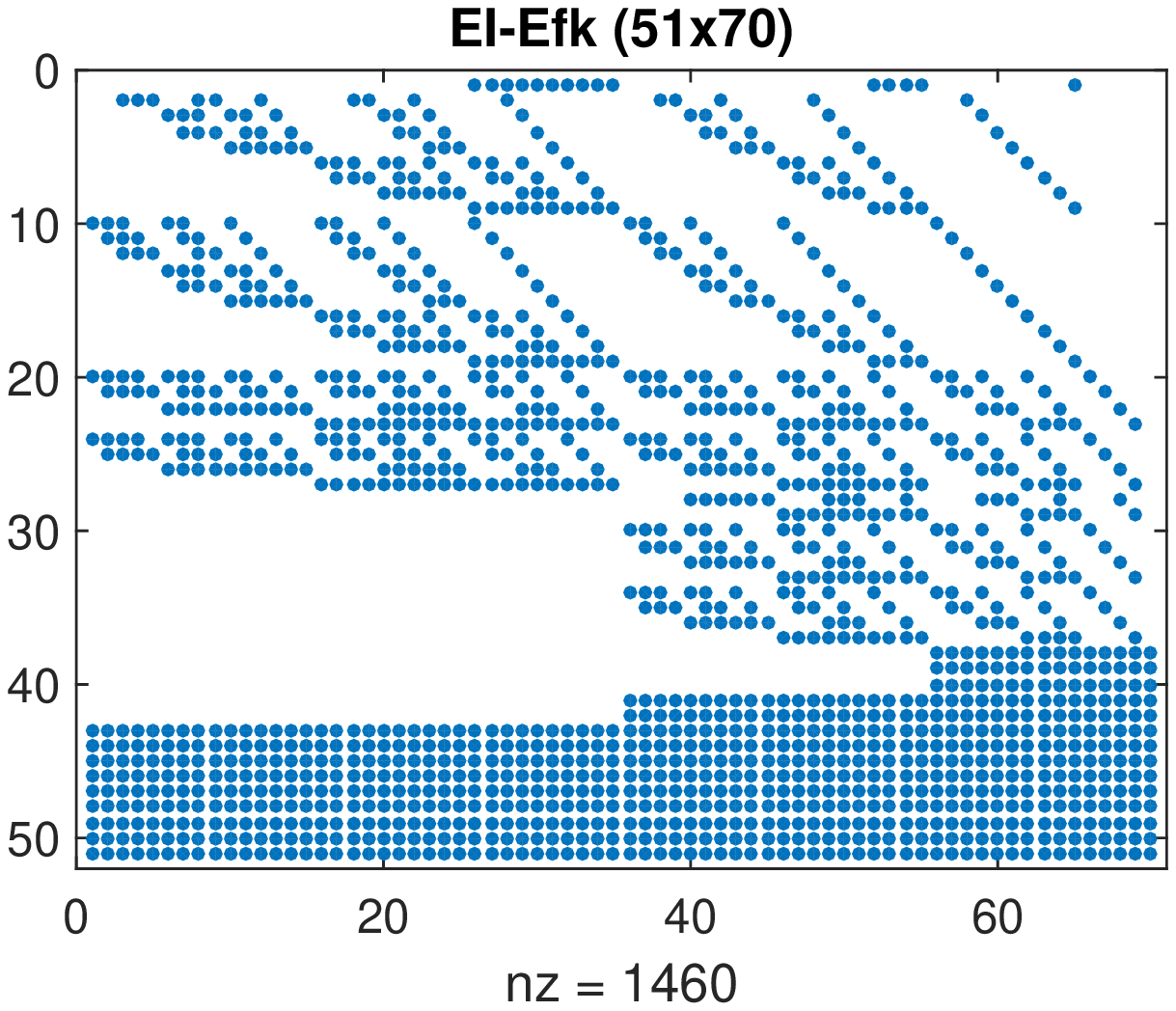}
\vspace{-0.2cm} 
\tabularnewline
\vspace{-0.0cm} 
{\scriptsize{(a)}} &
{\scriptsize{(b)}} 
\end{tabular}
\caption{{\bf Sparsity  patterns for the solvers to the \Efk problem:}  (a) state-of-the-art \ntm{200}{231} Kuang14~\cite{Kuang-CVPR-14} solver (b) the new \ntm{51}{70} EI-Efk solver.}
\label{fig:Efk_sparsity}
\end{figure}

Finally, Figure~\ref{fig:Efk_sparsity} shows the sparsity patterns of the (a) SOTA \ntm{200}{231} Kuang14~\cite{Kuang-CVPR-14} solver  and (b) the new \ntm{51}{70} EI-Efk solver for the \Efk problem.
Here, the ratio $nz_S /nz_{EI}$  is approximately 2.8.


%
%

\section*{Acknowledgement}
\noindent Z.~Kukelova was supported by The Czech Science Foundation Project GACR P103/12/G084. T.~Pajdla was supported by the EU-H2020 project LADIO No.~731970.
J.~Kileel and B.~Sturmfels were supported by
the US National Science Foundation (DMS-1419018)
and the Max-Planck Institute for Mathematics in the Sciences, Leipzig, Germany.

\bibliographystyle{unsrt}
\bibliography{CVPR-2017-ElimIdeal-arXiv.bib}
\end{document}